\documentclass[11pt]{article}
\usepackage[margin=1in]{geometry}

\usepackage{amsmath, amssymb, amsfonts}
\usepackage{graphicx}
\usepackage[section]{placeins} 
\usepackage{tikz-cd}
\usepackage{multicol}
\usepackage[utf8]{inputenc}
\usepackage[T1]{fontenc}

\usepackage{mathtools}
\usepackage{bm}
\usepackage{authblk}
\usepackage{hyperref}
\usepackage{xcolor}
\usepackage{soul}
\sethlcolor{yellow}
\usepackage{natbib}
\usepackage{caption} 
\usepackage{booktabs}
\usepackage{longtable}
\usepackage{array}
\usepackage{makecell}
\usepackage{enumitem}

\newcommand\blfootnote[1]{%
  \begingroup
  \renewcommand\thefootnote{}\footnotetext{#1}%
  \addtocounter{footnote}{-1}%
  \endgroup
}

\newcommand{\R}{\mathbb{R}}

\title{A Multimodal Foundation Model for Longitudinal Patient Representation and Scalable Insight Generation in Oncology}

\author{
Eugene Vorontsov\textsuperscript{*},
Yi Kan Wang\textsuperscript{*},
Alican Bozkurt\textsuperscript{*},
Adam Casson\textsuperscript{*},
Ludmila Tydlitátová\textsuperscript{*},
Michal Zelechowski\textsuperscript{*},
Ezra E. W. Cohen,
Jyoti D. Patel,
Max Banaszak,
Caitlin McWilliams,
Shane Colley,
Kate Sasser,
Ryan Fukushima,
Eric Lefkofsky,
Razik Yousfi,
Siqi Liu\textsuperscript{*,\textdagger}
}

\affil{Tempus AI, Inc., Chicago, IL, USA}

\date{}

\begin{document}

\maketitle
\noindent\textsuperscript{*}These authors contributed equally to this work.\\
\textsuperscript{\textdagger}Corresponding author.
\blfootnote{This is a living preprint describing preliminary work and will be updated as more analyses and results mature.}

\begin{abstract}
\end{abstract}
Precision oncology necessitates a longitudinal model of patient state that captures cancer evolution and treatment over time, integrating multimodal observations. We introduce the oFM, a foundation model developed on a real-world oncology cohort of 1.67 million cancer patients that integrates clinical trajectories with DNA, RNA, and H\&E pathology. Patient-level partitions were reserved for training, validation, and testing, with over one million patients used for training. The oFM encodes daily clinical and molecular episodes and, along with pathology images, integrates them over time to produce a patient state embedding. We evaluate frozen oFM embeddings against expert-curated clinical and molecular baseline features. In prognostic benchmarks, the oFM improved AUC for treatment response, progression-free survival, and overall survival (0.774 vs. 0.563 for overall survival). Across 11 comparative-treatment cohorts, the oFM embeddings achieved a three-fold higher pooled and scale-normalized treatment-benefit AUTOC than baseline features with improved benefit ranking in 9 of 11 cohorts, and provided stronger prognostic discrimination within both treatment arms. We also evaluated a mechanism discovery framework that interprets downstream models built on oFM embeddings by linking their predicted outcomes to clinically and biologically grounded mechanisms through an evidence-grounded temporal graph, enabling evaluation in clinical and drug-development applications.

\section{Introduction}
Oncology is a longitudinal study. Treatment must adapt to evolving patient states \citep{murphy2003optimal,chakraborty2014dynamic,strobl2023treatment}. However, routine clinical decision systems using clinical variables and biomarkers remain static snapshots that ignore the dynamic cancer behavior that drives metastatic potential, treatment response or resistance, and toxicity burden. Furthermore, while foundation models have revolutionized representation learning within individual biomedical modalities, including pathology \citep{vorontsov2024foundation,UNI2,Zimmermann2024Virchow2,Vorontsov2026PRISM2}, genomics \citep{brixi2026genome,theodoris2023geneformer}, clinical text \citep{Yang2022GatorTron}, and structured electronic health records (EHR) \citep{guo2024multicenter}, simply assembling modality-specific encoders fails to capture how biology and health evolve over time and across interventions. A foundation model for precision oncology must handle longitudinal data and support the development of predictive biomarkers, not only prognostic biomarkers. Predictive biomarkers quantify differential treatment benefit whereas prognostic biomarkers estimate risk independent of treatment and may identify patients with aggressive disease without revealing which treatment could alter their trajectory~\citep{ballman2015biomarker}.

\begin{figure}[!tbp]
\centering
\includegraphics[width=\linewidth,height=0.85\textheight,keepaspectratio]{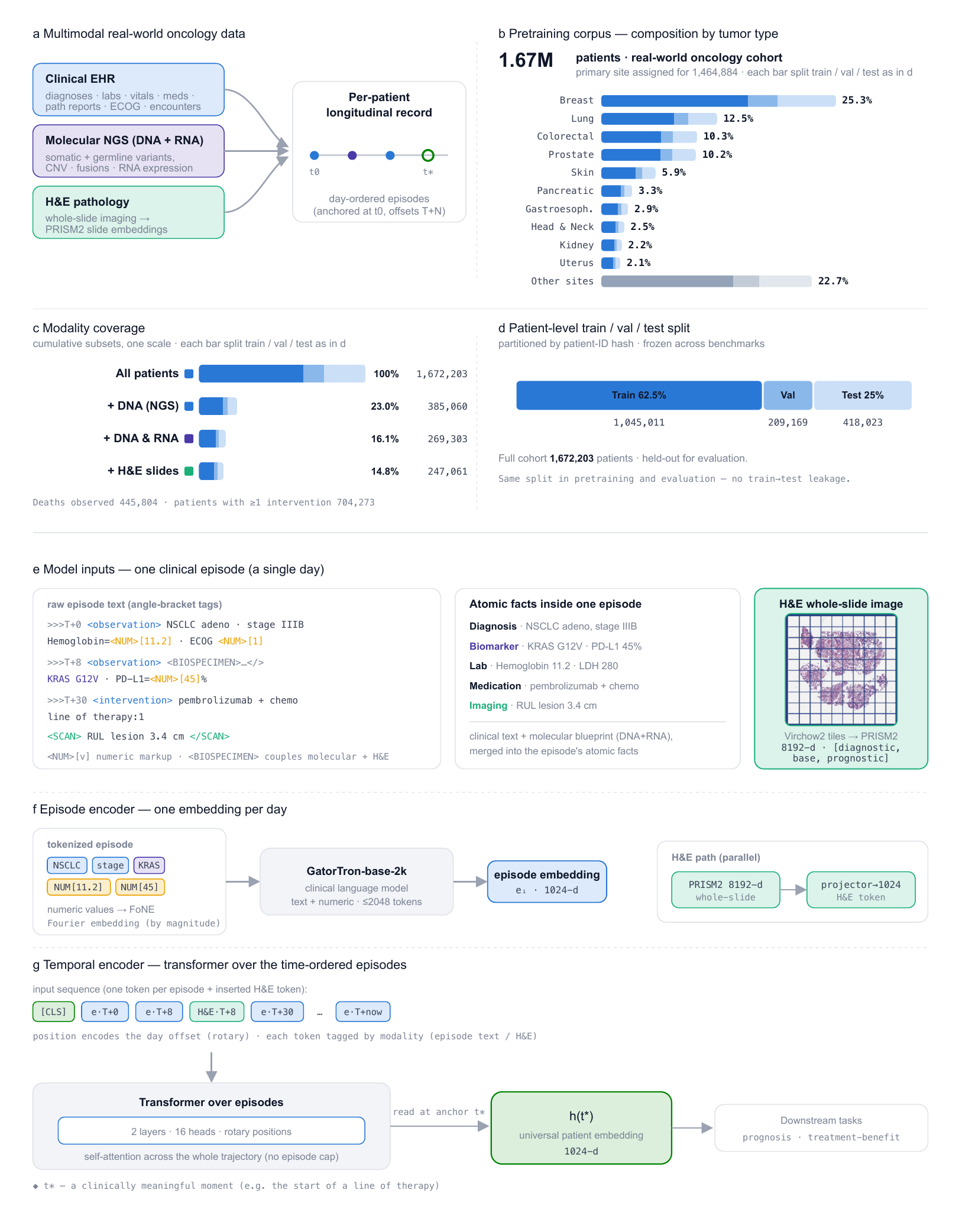}
\captionsetup{font=small}
\caption[oFM data and architecture.]{\textbf{oFM data and architecture.} \textbf{(a)} Longitudinal patient records integrate clinical records (EHR and concepts curated from free-text reports), molecular NGS, IHC and ISH, and H\&E pathology data. \textbf{(b-d)} The pretraining cohort is summarized by tumor type, modality coverage, and patient-level data partition. \textbf{(e-f)} Daily clinical and molecular facts are encoded as episode embeddings, while pathology embeddings are projected into the shared latent space. \textbf{(g)} A temporal Transformer integrates date-stamped, modality-tagged tokens to produce the patient representation $h(t^*)$ for downstream prognostic and treatment-benefit modeling.}
\label{fig:ofm-overview}
\end{figure}

We introduce the oFM, a longitudinal multimodal foundation model for oncology (Figure~\ref{fig:ofm-overview}). The oFM integrates Tempus-curated longitudinal records with patient-linked molecular profiles and H\&E pathology embeddings. Diverse clinical and molecular events such as laboratory measurements, diagnoses, procedures, treatments, and DNA and RNA biomarkers, are decomposed into atomic facts. Facts are aggregated into daily episodes, embedded with a finetuned language model. A transformer-based trajectory encoder then integrates these episode-level and pathology embeddings to produce a patient trajectory embedding. These embeddings support multimodal patient subtyping, novel biomarker discovery, and scalable AI-enabled diagnostic development. We show that oFM embeddings significantly improve prognostic prediction and the prediction of treatment benefit over traditional clinical features and biomarkers in real-world comparative cohorts. Finally, we explore a mechanism-discovery framework that decomposes model predictions into attributed clinical and molecular features, sparse latent concepts, and evidence-grounded temporal relationships. By combining latent-space intervention with retrieval from biomedical knowledge sources, the framework organizes these signals into an interpretable temporal graph, providing additional oncology insights.

\section{Related Work}

\paragraph{Multimodal Cancer Survival Models}
MultiSurv~\citep{ValeSilva2021MultiSurv}, MCAT~\citep{Chen2021MCAT}, and PORPOISE~\citep{Chen2022PORPOISE} integrate clinical, histopathologic, and molecular data at a single time point. In contrast, the oFM models multimodal observations longitudinally capturing changing patient trajectories.

\paragraph{Longitudinal patient and EHR foundation models.}
BEHRT and Med-BERT use masked language modeling on sequences of structured EHR codes~\citep{Li2020BEHRT,Rasmy2021MedBERT}, while CEHR-BERT introduced time encoding between events~\citep{Pang2021CEHRBERT}. CLMBR, ETHOS, Foresight, and CoMET used an autoregressive objective which implies time progression~\citep{Steinberg2020CLMBR,Renc2024ETHOS,Kraljevic2024Foresight,Waxler2025CoMET}. Delphi-2M and MOTOR introduced time-to-event objectives~\citep{Shmatko2025Delphi2M,Steinberg2023MOTOR}, while SMB-v1 predicts future latent states~\citep{Adam2026SMB}. Beyond structured EHR data, APOLLO integrates unstructured clinical text embeddings and medical image embeddings covering whole-slide H\&E pathology images, hematology blood smears, and electron microscopy~\citep{Zhang2026APOLLO}. Whereas APOLLO uses masked language modeling over patient trajectories, the oFM handles future prediction via treatment-conditioned latent space prediction and demonstrates oncology-focused treatment benefit stratification on real world data.

\paragraph{Clinical-text and pathology foundation encoders.}
The oFM uses GatorTron~\citep{Yang2022GatorTron} to encode clinical and molecular text and PRISM2~\citep{Vorontsov2026PRISM2} to encode slide-level H\&E embeddings. This modular design allows the oFM to benefit from continued improvements in modality encoders.

\paragraph{Combining pathology and molecular modalities.}
THREADS~\citep{threads2025} and mSTAR~\citep{mstar2025} combine pathology images with molecular signal for a single specimen. OmniScreen predicts DNA biomarkers directly from H\&E images \citep{screenThemAll2024}. However, these models operate at a single time point and do not support longitudinal patient trajectories.

\paragraph{Genomic foundation models.}
Sequence and cancer-genomics foundation models such as DNABERT, DNABERT-2 \citep{dnabert2021,dnabert2_2024}, Nucleotide Transformer \citep{nucleotidetransformer2023}, HyenaDNA \citep{hyenadna2023}, and Enformer \citep{enformer2021} learn reusable molecular representations. Related single-cell models such as Geneformer and scGPT provide complementary representations at the expression level \citep{theodoris2023geneformer,scgpt2024}. The oFM integrates molecular inputs over time by ingesting variant and expression summary text alongside other clinical and pathology inputs. The choice of alternative molecular input encoding in the oFM will be explored in further work.

\section{The oFM Model Architecture}

The oncology foundation model (oFM) is a multimodal Transformer that models a cancer patient's evolving clinical state. Given patient history observed up to time $t^*$, the oFM produces a patient embedding $h(t^*) \in \R^{1024}$. The hierarchical architecture uses an episode encoder to summarize clinical and molecular information in a day as an episode and a temporal encoder to integrate these episode and pathology embeddings over time. See Figure~\ref{fig:ofm-overview} for an overview.

\subsection{Episode Encoder}

An episode contains Tempus-curated clinical facts and molecular measurements (see Appendix~\ref{appendix:data_source} for data curation) attributed to a single day. One or more facts form a time-stamped event. Events are indexed by the number of elapsed days from the first observed event in the patient trajectory. Demographic information forms a separate episode at the beginning of the patient trajectory. Clinical facts include patient demographics, diagnoses, metastases, laboratory measurements, vital signs, performance status, receptor status, treatments, procedures, responses, progression, and follow-up events. Molecular findings per biospecimen include DNA alterations, copy-number variants, fusions, RNA expression abnormalities, splice events, and IHC or NGS biomarkers. All clinical and molecular events are input as text to the episode encoder.

The episode encoder is a finetuned GatorTron-base-2k, a clinical BERT model with hidden dimension 1,024 and a maximum input length of 2,048 tokens \citep{Yang2022GatorTron}. The final-layer CLS representation is used as the episode embedding. To preserve measurement magnitudes, we augment the language encoder with a Fourier Number Embedding (FoNE) pathway \citep{Zhou2026FoNE}. Each value is first transformed using its signed log-magnitude and then represented using sine and cosine features at 32 frequencies spanning 0.1 to 10.0. A learned multi-layer perceptron (MLP) and layer normalization (LayerNorm) project these features into the 1,024-dimensional token space. The resulting representation is added to the token embedding at the corresponding \texttt{[NUM]} position, enabling the episode encoder to model both semantic context and magnitude.

H\&E whole slide images (WSIs) are embedded alongside episode embeddings as one embedding per biospecimen. We tokenize each WSI with Virchow2~\citep{Zimmermann2024Virchow2} and produce a single biospecimen embedding of one or more WSIs with PRISM2 \citep{Vorontsov2026PRISM2}. This single embedding is a concatenation of the PRISM2 Base, Diagnostic, and Survival embeddings into a single 8,192-dimensional vector. To map this vector into the trajectory latent space, we train a LayerNorm and linear projection mapping from 8,192 to 1,024 dimensions.

\subsection{Trajectory Encoder}

The episode and pathology embeddings are aggregated by a Transformer-based multimodal trajectory encoder in a shared 1,024-dimensional latent space. The encoder consists of two Transformer layers (16 attention heads, 4,096-dimensional feed-forward layer, dropout probability 0.1, and FlashAttention). A learned modality encoding $\mathbf{m}_k \in \R^{1024}$, $k \in \{\mathrm{e}, \mathrm{p}\}$ is added to each token to distinguish episode embeddings from pathology embeddings. Relative time is represented using rotary positional embeddings (RoPE) \citep{Su2021RoPE}. A single day may contain up to one episode token and any number of pathology tokens. Thus a day $i$ containing both an episode embedding $\mathbf{x}_{\mathrm{e}}$ and a biospecimen-level embedding $\mathbf{x}_{\mathrm{p}}$ of pathology whole slide images produces a set of tokens:

\begin{equation}
\mathbf{d}_i =
\left[
\mathbf{x}_{\mathrm{e}} + \mathbf{m}_{\mathrm{e}},
\mathbf{x}_{\mathrm{p}} + \mathbf{m}_{\mathrm{p}}
\right].
\end{equation}

For a sequence of $D$ observed days, the trajectory encoder observes as a sequence:

\begin{equation}
\mathcal{H}_D = 
\left[
\mathbf{c},
\mathbf{d}_1,
\ldots,
\mathbf{d}_D
\right],
\end{equation}

where $\mathbf{c} \in \R^{1024}$ is a learned CLS token. When there is no observed input for a calendar day, it is excluded from the sequence. All tokens attend to each other bidirectionally. The output CLS token of the trajectory encoder is used as the trajectory embedding $\mathbf{h}$ at time $t^*$:

\begin{equation}
\mathbf{h}(t^*)
=
\operatorname{TemporalEncoder}_{\mathrm{CLS}}
\left(
\mathcal{H}_{\tau}
:
\tau \le t^*
\right)
\in \R^{1024}.
\end{equation}

Evaluating $\mathbf{h}$ at different clinical anchor points enables longitudinal comparison.

\section{Training the oFM}

\begin{figure}[!htb]
  \centering
  \includegraphics[width=\linewidth]{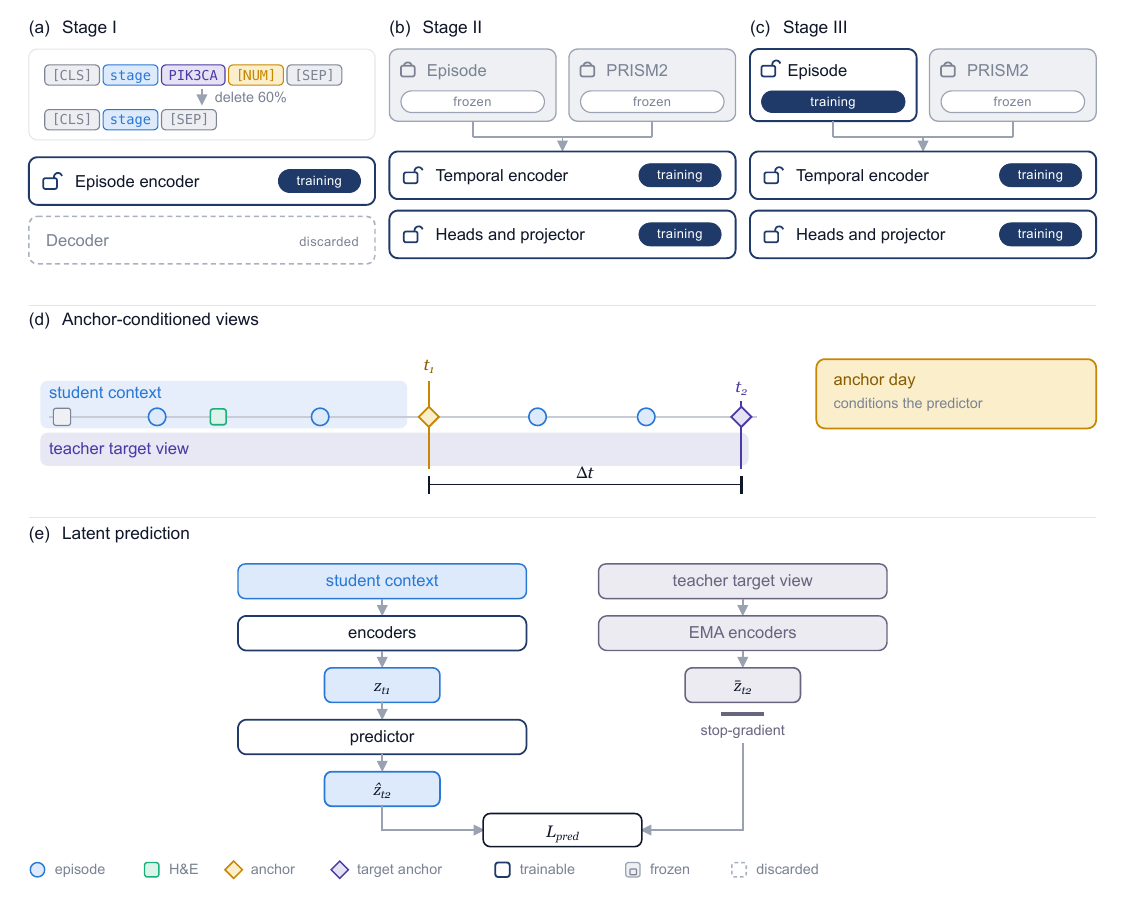}
  \caption{\textbf{oFM training curriculum.} \textbf{(a)} Stage I pretrains the episode encoder by reconstructing missing tokens in per-day episodes. \textbf{(b)} Stage II freezes the episode and PRISM2 encoders while training the temporal encoder, prediction heads, and projectors. \textbf{(c)} Stage III unfreezes the episode encoder. \textbf{(d)} Anchor-conditioned training compares a student view to a later teacher view, with prediction conditioned on the anchor-day intervention and elapsed time. \textbf{(e)} The student predicts the stop-gradient representation produced by the exponential-moving-average teacher in latent space.}
  \label{fig:ofm-training}
\end{figure}

The oFM is trained using a three-stage curriculum (Figure~\ref{fig:ofm-training}). Stage I pretrains the episode encoder using token- and numerical-reconstruction objectives. Stage II freezes the episode encoder and trains the trajectory encoder. Stage III unfreezes the episode encoder for joint end-to-end training. Stage II and III use full patient trajectory inputs and combine masked episode reconstruction, self-supervised future latent prediction, and a supervised survival objective that encourages the patient trajectory representation to retain outcome-relevant information.

Stage I pretrains the episode encoder on individual episodes using Transformer-based Sequential Denoising Auto-Encoding (TSDAE) \citep{Wang2021TSDAE}. For each episode, input tokens (excluding \texttt{[CLS]} and \texttt{[SEP]}) are randomly dropped with a $60\%$ probability and the model is trained to reconstruct the complete uncorrupted episode from the output CLS token. This reconstruction is performed by an autoregressive Transformer-based decoder that shares the same depth, self-attention, feed-forward, and layer-normalization parameters, while adding cross-attention to the CLS token. FoNE is trained jointly for numerical values. In addition to token reconstruction, a decoder-side regression head reconstructs numerical values from the decoder hidden states at \texttt{[NUM]} positions. Numerical values $v$ are transformed using the signed log-magnitude, $\operatorname{sign}(v)\log(1+|v|)$, and regressed with a smooth-L1 loss. The complete Stage I objective is

\begin{equation}
\mathcal{L}_{\mathrm{Stage\,I}}
=
\mathcal{L}_{\mathrm{token}}
+
\mathcal{L}_{\mathrm{num}},
\end{equation}

where $\mathcal{L}_{\mathrm{token}}$ is the autoregressive token-reconstruction loss and $\mathcal{L}_{\mathrm{num}}$ is the numerical reconstruction loss. This pretrains the episode encoder, tokenizer, and FoNE while the decoder is discarded.

Stages II and III train the trajectory encoder and restrict the training cohort to patients with recorded treatment information. Episodes are grouped into whole per-patient trajectory inputs. The pretrained episode encoder is kept frozen in Stage II and is unfrozen in stage III. The linear projection of episode and pathology embeddings to the trajectory encoder input is trained in both stages. Input trajectories are truncated at strategically sampled anchor points and training is supervised by masked episode reconstruction, joint-embedding trajectory prediction (JEPA), and survival risk supervision.

\paragraph{Masked episode reconstruction.}
Each episode or linearly projected pathology embedding is independently randomly masked with probability 0.7 by replacing its content representation with a learned episode- or pathology-specific mask embedding, retaining its temporal encoding. The trajectory encoder reconstructs the masked episodes and full pre-projection masked pathology embeddings using modality-specific reconstruction heads. Reconstruction is optimized with the mean squared error loss and, in the case of pathology embeddings, a cosine distance loss. The masked reconstruction loss is thus:

\begin{equation}
\begin{aligned}
\mathcal{L}_{\mathrm{rec}}
&= \mathcal{L}_{\mathrm{e}} + \mathcal{L}_{\mathrm{p}}
\\
\mathcal{L}_{\mathrm{e}}
&= \frac{1}{D_{\mathrm{e}}\lvert\mathcal{M}_{\mathrm{e}}\rvert}
   \sum_{i\in\mathcal{M}_{\mathrm{e}}}
   \bigl\lVert f_{\mathrm{e}}(\mathbf{z}_i)-\mathbf{x}_{\mathrm{e},i}\bigr\rVert_2^2
\\
\mathcal{L}_{\mathrm{p}}
&= 
\frac{1}{D_{\mathrm{p}}\lvert\mathcal{M}_{\mathrm{p}}\rvert}
\sum_{i\in\mathcal{M}_{\mathrm{p}}}
   \bigl\lVert f_{\mathrm{p}}(\mathbf{z}_i)-\mathbf{x}_{\mathrm{p},i}\bigr\rVert_2^2
   \;+\;
   \frac{\lambda_{\cos}}{\lvert\mathcal{M}_{\mathrm{p}}\rvert}
   \sum_{i\in\mathcal{M}_{\mathrm{p}}}
   \Bigl(1-\cos\bigl(f_{\mathrm{p}}(\mathbf{z}_i),\,\mathbf{x}_{\mathrm{p},i}\bigr)\Bigr)
\end{aligned}
\end{equation}

where $\mathcal{M}_{\mathrm{e}}$ and $\mathcal{M}_{\mathrm{p}}$ are the sets of masked episode and pathology token indices, $\mathbf{z}_i\in\mathbb{R}^{1024}$ is the trajectory encoder's output for masked token $i$, $\mathbf{x}_{\mathrm{e},i}\in\mathbb{R}^{D_{\mathrm{e}}}$ ($D_{\mathrm{e}}=1024$) is the unmasked episode embedding, $\mathbf{x}_{\mathrm{p},i}\in\mathbb{R}^{D_{\mathrm{p}}}$ ($D_{\mathrm{p}}=8192$) is the pre-projection pathology embedding, $f_{\mathrm{e}}$ and $f_{\mathrm{p}}$ are the modality-specific heads, and $\lambda_{\cos}=1$.

\paragraph{Anchor pair sampling.}
For each patient trajectory containing a sequence of episodes, we identify anchor episodes and sample a pair of anchors $(t_1, t_2)$ to predict outcome or patient state at $t_1$ given a patient state near $t_1$. An episode that contains an \texttt{<intervention>} line (treatment start, treatment end, or surgery) or a \texttt{<transition>} line (metastasis, progression, response assessment, last-known vital status) acts as an anchor. In a random pair, $t_1$ contains an intervention and $t_2$ is any intervention or transition after $t_1$.

\paragraph{Joint-embedding trajectory prediction.}
Following I-JEPA, we use a student-teacher setup with a predictor network that predicts a future patient state, as encoded by the teacher, from the student's corrupted patient state~\citep{Assran2023IJEPA}. Let $g_{\theta}$ and $h_{\theta}$ denote the student episode and trajectory encoders with parameters $\theta$ and write $f_{\theta}=h_{\theta}\circ g_{\theta}$; the teacher $f_{\bar{\theta}}$ is an exponential moving average of the student with parameters $\bar{\theta}$. The student and the teacher see two different views of the same trajectory, determined by the sampled anchor pair $(t_1,t_2)$. Because $t_1$ includes an intervention, the student context $\mathcal{H}_{<t_1}$ contains all episode and pathology embeddings strictly before the anchor day $t_1$ to ensure that no information about the intervention leaks into the context. Instead, the episode embedding on the day of the intervention ($\mathbf{x}_{\mathrm{e},t_1}$) conditions the predicted future state at $t_2$. The teacher target view $\mathcal{H}^{\mathrm{tgt}}_{\le t_2}$ extends the trajectory through the target anchor: it contains all episode and pathology embeddings up to and including day $t_2$ when $t_2$ is a \texttt{<transition>} anchor, so that the target representation reflects the observed outcome, and up to but not including day $t_2$ when $t_2$ is an \texttt{<intervention>} anchor, applying the same no-leak rule as at $t_1$. The residual predictor $\Delta_\phi$ thus maps the student's corrupted context state to the teacher's future patient state:

\begin{equation}
\begin{aligned}
\mathbf{z}_{t_1} &= f_{\theta}\!\left(\widetilde{\mathcal{H}}_{<t_1}\right), \\
\bar{\mathbf{z}}_{t_2} &= f_{\bar{\theta}}\!\left(\mathcal{H}^{\mathrm{tgt}}_{\le t_2}\right), \\
\widehat{\mathbf{z}}_{t_2} &= \mathbf{z}_{t_1}
  + \Delta_{\phi}\!\left(\mathbf{z}_{t_1},\, \mathbf{x}_{\mathrm{e},t_1},\, \psi(\Delta t)\right),
\end{aligned}
\end{equation}

where all three vectors lie in $\mathbb{R}^{1024}$, $\Delta t = t_2 - t_1$, and $\psi$ is a sinusoidal embedding of elapsed time refined by a two-layer MLP. $\Delta_{\phi}$ is a two-layer MLP whose input and hidden activations are FiLM-modulated by $[\mathbf{x}_{\mathrm{e},t_1};\psi(\Delta t)]$~\citep{Perez2018FiLM}.
The teacher target view is encoded without masking. The corrupted student context $\widetilde{\mathcal{H}}_{<t_1}$ is obtained from $\mathcal{H}_{<t_1}$ using the same mask as the masked episode reconstruction objective: each token is replaced by the learned modality-specific mask embedding with probability $0.7$.

The predictive loss is $\mathcal{L}_{\mathrm{pred}}=\tfrac{1}{1024}\bigl\lVert\widehat{\mathbf{z}}_{t_2}-\operatorname{sg}[\bar{\mathbf{z}}_{t_2}]\bigr\rVert_2^{2}$, averaged over the minibatch, with $\operatorname{sg}[\cdot]$ the stop-gradient. The teacher receives no gradient and is updated after each optimizer step as $\bar{\theta}\leftarrow m\bar{\theta}+(1-m)\theta$ with momentum $m=0.999$, following BYOL and I-JEPA \citep{Grill2020BYOL,Assran2023IJEPA}. Both the $\Delta_{\phi}$ state predictor and the teacher are discarded after training.

To prevent representational collapse we add VICReg's variance and covariance terms ($\mathcal{L}_{\mathrm{var}}$ and $\mathcal{L}_{\mathrm{cov}}$) on $\widehat{\mathbf{z}}_{t_2}$ \citep{Bardes2022VICReg}: a hinge holding each dimension's standard deviation at or above $\gamma=1$, weighted $\lambda_{\mathrm{var}}=1$, and the mean squared off-diagonal covariance, weighted $\lambda_{\mathrm{cov}}=0.04$. 

\paragraph{Pathology dropout.} To improve robustness to missing pathology data, each pathology embedding is randomly removed with probability 0.3 during training. A removed embedding is dropped simultaneously for both the student and the teacher networks. Dropped embeddings are excluded entirely from the trajectory and are not treated as masked-reconstruction targets.

\paragraph{Survival supervision.}
Trajectory training is supervised by overall survival prediction. A survival head predicts a risk score from the patient state embedding $h(\mathcal{H}_{<t_1})$, up to but not including the anchor day $t_1$, concatenated with the episode embedding $\mathbf{x}_{\mathrm{e},t_1}$ at $t_1$. Survival time is measured from the anchor day to death or to the last recorded event (right-censoring), and the model is optimized using the Cox partial-likelihood loss $\mathcal{L}_\mathrm{Cox}$ with Efron handling of tied event times. Risk sets are constructed over the globally gathered distributed minibatch.

The complete trajectory-level objective in stages II and III is:

\begin{equation}
\mathcal{L}_{\mathrm{trajectory}}
=
\mathcal{L}_{\mathrm{rec}}
+
\mathcal{L}_{\mathrm{pred}}
+
\mathcal{L}_{\mathrm{var}}
+
0.04\,\mathcal{L}_{\mathrm{cov}}
+
\mathcal{L}_{\mathrm{Cox}},
\end{equation}

with all loss terms weighted equally except the covariance regularizer.

Both stages are optimized with AdamW ($\beta_1=0.9$, $\beta_2=0.999$) at bfloat16 mixed precision with gradient clipping at a maximum norm of 1.0. Trajectory components (trajectory encoder, predictor, time embedding, projection and reconstruction heads, and survival head) train at a learning rate of $10^{-4}$ with weight decay 0.01. A 1000-step linear warmup is used for the trajectory encoder. Stage III unfreezes the pretrained episode encoder and trains it with a $10^{-4}$ learning rate with weight decay 0.01 ($10^{-5}$ for the token-embedding matrix, without weight decay).

\section{Data}
\label{sec:data}

The oFM was developed within a 1,672,203 patient cohort of the Tempus de-identified multimodal real-world oncology corpus (Fig. \ref{fig:ofm-overview}). Of these, 1,045,011 patients were used for training the episode encoder (stage I), a subset of 386,382 patients (109,938 with pathology H\&E images), all with recorded treatment events, were used to train the temporal encoder (stage II), and a subset of 92,567 information-rich patients with both paired DNA and RNA records were used for final end-to-end finetuning (stage III). Patient records were structured as longitudinal sequences anchored at the earliest observed event, indexing subsequent events by elapsed calendar days to preserve irregular clinical intervals. All test set patients in evaluation tasks were excluded from model training. The data used to train and evaluate the oFM is distinct and non-overlapping with the data used to train Virchow2 and PRISM2.

The longitudinal record integrates clinical events, molecular blueprints, and digitized H\&E pathology over time. Clinical information spans demographics, encounters, vitals, labs, diagnoses, treatments, procedures, and outcomes. These data are organized into observations, interventions, and disease-state transitions. Observations describe measured or recorded patient characteristics. Interventions include treatments and procedures. Transitions represent clinically relevant changes such as new metastasis, tumor progression, treatment response, or last-known follow-up. Molecular findings for each tumor biospecimen were consolidated into a standardized textual record termed a \emph{molecular blueprint}. These include DNA sequencing, whole-transcriptome RNA sequencing, immunohistochemistry (IHC), and in situ hybridization (ISH). See Appendix~\ref{appendix:blueprint} for more details.

\section{Downstream Evaluation}

The downstream benchmarks evaluate the relative quality of the learned patient trajectory embeddings. Despite the challenges related to retrospective real-world data (e.g., confounding, censoring), we demonstrate that the oFM provides a prognostic and predictive advantage over a strong task-specific baseline. Specifically, there is a substantial and statistically significant improvement in aggregate performance across benchmarks.

In comparison to the oFM embeddings, we evaluate a strong curated-feature baseline that selects from 7520 candidate curated clinical and molecular features (NGS: DNA and RNA; IHC) using the feature selection process described in Appendix~\ref{appendix:feature-selection}. All prognostic and predictive evaluation was performed using the two-stage convex UV-Cox method presented in Appendix~\ref{sec:uv-optimization}.

\subsection{Prognostic Classification}

We perform binary linear probing on three endpoint measures to evaluate outcome-prognostic performance of oFM embeddings encoded just prior to the initiation of first-line therapy:

\begin{enumerate}[label=(\roman*)]
\item \textbf{Treatment response groups:} Responder (Complete or Partial, CR/PR) versus others (stable or progressive, SD/PD), using the latest assessment between 14 and 365 days after first-line treatment initiation and prior to the second line.

\item \textbf{Progression-free survival (PFS) groups:} Sustained progression-free status vs early progression. 
Positive if PFS above 60th percentile; negative if below 30th percentile.

\item \textbf{Overall survival (OS) groups:} Long-term vs short-term survival. Positive OS above 60th percentile; negative if below 30th percentile.
\end{enumerate}

For each task, we train a logistic-regression probe on frozen patient embeddings extracted just before first-line initiation (see Appendix~\ref{sec:uv-optimization} for method). To reduce prognostic confounding by tumor type and therapy class, we stratify test-set performance by tumor type $\times$ therapy-class into 39 / 38 / 40 strata for OS / PFS / treatment-response, retaining at least 20 patients per stratum, and compute the mean (see Appendix Tab.~\ref{tab:app-per-stratum-composition} for strata).

\begin{figure}[htbp]
\centering
\includegraphics[width=\linewidth]{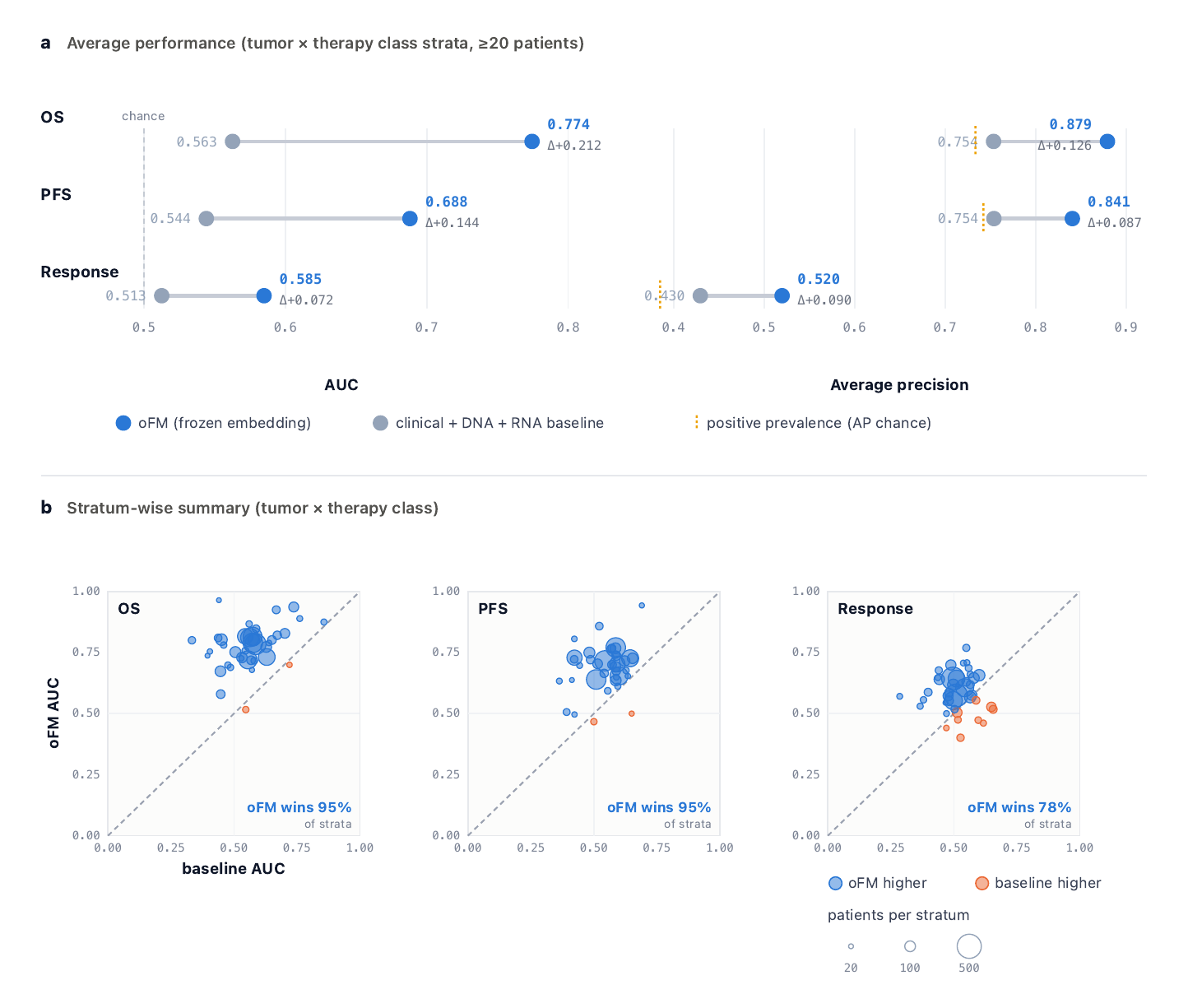}
\caption{\textbf{Prognostic classification (tumor $\times$ therapy stratified).} \textbf{(a)} Stratified summary (macro-mean AUC and AP) for response, PFS group, and OS group, comparing a frozen-embedding probe (oFM) to an engineered-feature baseline under an identical probe. \textbf{(b)} Per-stratum AUC comparison (one point per tumor $\times$ therapy stratum; point area proportional to patients) showing that gains are broad-based and concentrated in larger strata.}
\label{fig:app-prognostic-cls}
\end{figure}

As shown in Fig.~\ref{fig:app-prognostic-cls}, linear probing on the oFM embeddings outperforms baseline features on all three endpoints, achieving 0.774 mean AUC for OS, 0.688 for PFS, and 0.585 for treatment response (CR/PR vs SD/PD), compared with 0.563, 0.544, and 0.513 for the baseline, respectively. The oFM performance beat the baseline on 95\% of OS and PFS strata and 78\% of treatment response strata.

\subsection{Predictive Benchmarks: Treatment Benefit}

\label{sec:cohorts}
\label{sec:predictive-survival}

We evaluate the oFM on benchmark cohorts spanning breast, colorectal, non-small-cell lung (NSCLC), renal-cell, prostate, and pan-cancer disease settings via Cox-based linear probing (using the UV-Cox method presented in Appendix~\ref{sec:uv-optimization}). Each benchmark poses a clinical question noted per cohort in Appendix Tab.~\ref{tab:cohorts-pred}. In the predictive setting, two treatment arms are compared in each cohort for differential treatment benefit. We use inverse-propensity weighting for predictive benchmarking to reduce data imbalance across arms inherent in real-world data (see Appendix~\ref{sec:uv-ipw}).

As shown in Fig.~\ref{fig:surv-predictive}, the oFM ranks treatment benefit and discriminates the baseline risk better than the curated-feature baseline. Fig.~\ref{fig:surv-predictive}a reports the $t_{\mathrm{AUTOC/SD}}$, the pooled AUTOC/SD normalized by $\mathrm{SE}_{\mathrm{AUTOC}}$, which measures how consistently nonzero the treatment benefit effect is across cohorts, and the pooled HRR (Paule-Mandel random effects; metrics detailed in Appendix~\ref{sec:uv-metrics}). The oFM achieves significant (p = 0.0003) treatment benefit ranking at $t_{\mathrm{AUTOC/SD}} = 4.61$ compared to the baseline at 1.38 which does not clear its permutation null of 1.85 (see Appendix~\ref{sec:uv-metrics} for definition); while the baseline does extract treatment benefit signal, it is not a consistently large effect. Fig.~\ref{fig:surv-predictive}b shows per-cohort wins: the oFM wins over the baseline in 9 of 11 cohorts.

\begin{figure}[!htb]
\centering
\includegraphics[width=\linewidth]{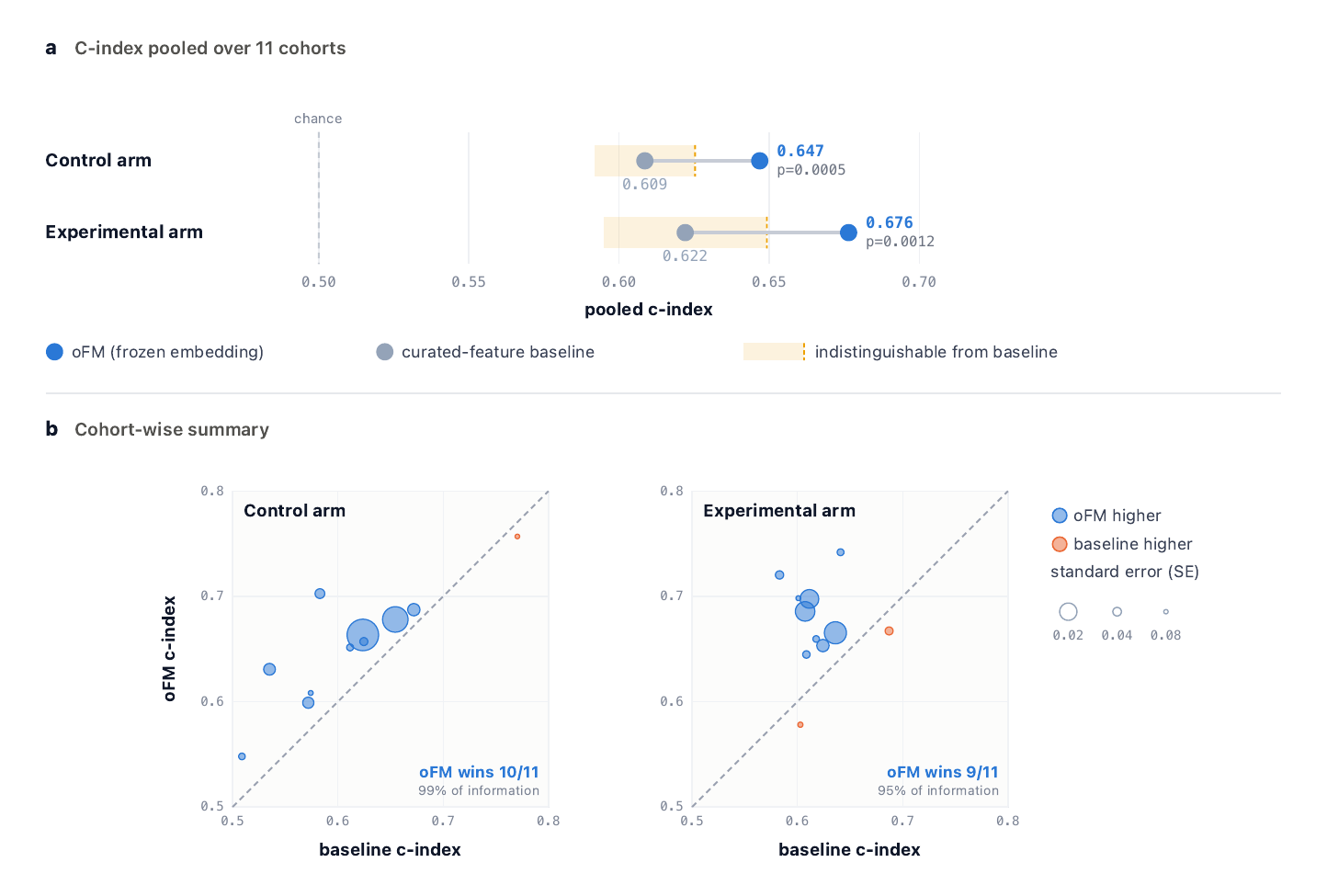}
\caption{\textbf{Prognostic discrimination.} Uno's C-index computed within the control arm (left) and the experimental arm (right), for oFM against the curated-feature baseline. \textbf{(a)} Pooled across cohorts by Paule--Mandel random effects: baseline C-index (grey) and oFM (blue) with $p$ the significance of the difference via two-sided Student t-test. The yellow band shows the range indistinguishable from the baseline. \textbf{(b)} One point per cohort and arm for baseline or oFM, whichever scores better; area proportional to $1/\mathrm{SE}^2$. Points above the diagonal favor oFM.}
\label{fig:surv-uno-cindex}
\end{figure}

Fig.~\ref{fig:surv-uno-cindex} shows per-arm prognostic performance. As shown in Fig.~\ref{fig:surv-uno-cindex}a, the  oFM embeddings produce better prognostic risk ordering (Uno's IPCW C-index, pooled with Paule-Mandel random effects) per arm than baseline features. In Fig.~\ref{fig:surv-uno-cindex}b, the oFM outperforms the baseline on 10 and 9 of 11 tasks in control and experimental arms, respectively. That the gain in C-index is similar in both the control and experimental arms is evidence that the signal is prognostic rather than a treatment effect leaking into the score.

\begin{figure}[!htb]
\centering
\includegraphics[width=\linewidth]{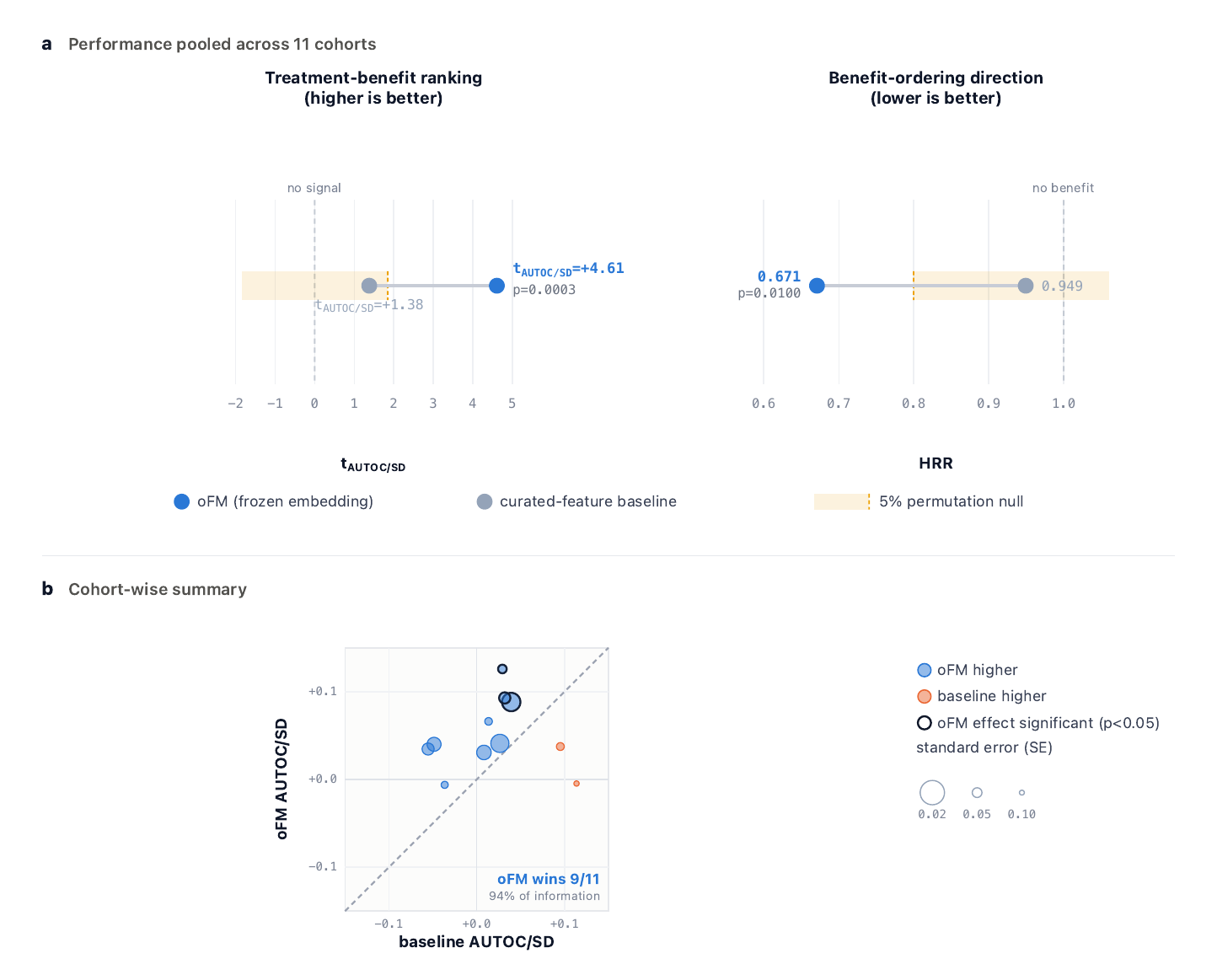}
\caption{\textbf{Predictive evaluation.} oFM frozen embeddings against the curated-feature baseline. \textbf{(a)} Metrics pooled across tasks by Paule-Mandel random effects: (left) $t_{\mathrm{AUTOC/SD}}$ is the pooled AUTOC/SD normalized by $\mathrm{SE}_{\mathrm{AUTOC}}$ and measures how consistently nonzero the treatment benefit effect is across cohorts and (right) the pooled HRR. The yellow bands are the 5\% permutation nulls, about zero (no signal) for the AUTOC/SD ranking and about one (no benefit) for the HRR; p values are computed with respect to each null. \textbf{(b)} Per-cohort AUTOC/SD, oFM against baseline. Area is proportional to $1/\mathrm{SE}^2$; points above the diagonal favor oFM and a bold outline marks $p < 0.05$.}
\label{fig:surv-predictive}
\end{figure}

\section{Mechanism Discovery}
\label{sec:mechanism}

Beyond measuring predictive performance, we seek to make downstream models built on oFM embeddings biologically and clinically interpretable by identifying human-interpretable mechanisms that drive their predictions. Mechanism discovery relies on finding sparse directions in embedding space that: (i)~are differentially active between high- and low-scored patients, (ii)~semantically map to external clinical and biological concepts, (iii)~causally alter downstream predictions under activation steering, and (iv)~assemble into a temporally ordered graph connecting concepts to outcomes via statistically tested precedence. Because the oFM jointly represents longitudinal clinical history with patient-linked DNA and RNA profiles, this framework enables downstream predictions to be interrogated in terms of observed clinical and tumor-biological factors, including molecular mechanisms associated with prognosis, treatment response, and resistance. These mechanisms can then be evaluated in downstream clinical and drug-development applications. The mechanism discovery pipeline is applied per evaluation cohort on frozen oFM embeddings $h(t^*)$ and a fitted predictor $s(h)$ in seven stages (Figure~\ref{fig:mech-pipeline}):

\begin{figure*}[t]
\centering
\includegraphics[width=\textwidth]{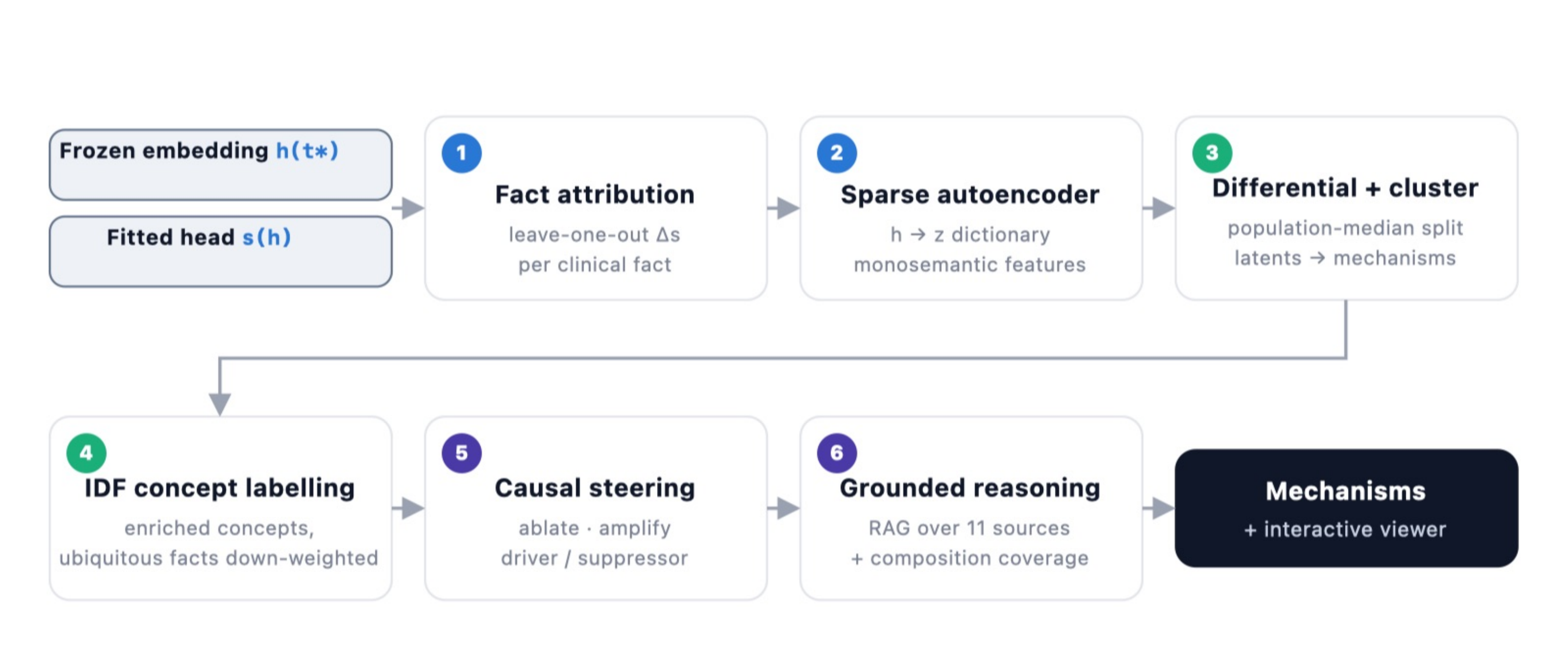}
\caption{\textbf{Mechanism-discovery pipeline.} Seven steps decompose a frozen
embedding $h(t^{*})$ and fitted head $s(h)$ into interpretable mechanisms:
(1)~leave-one-out fact attribution; (2)~sparse-autoencoder dictionary learning;
(3)~differential activation and clustering; (4)~IDF-enriched concept labelling;
(5)~causal steering via latent-space ablation/amplification; (6)~composition analysis
and retrieval-grounded reasoning; (7)~temporal-precedence mechanism graph with curated
knowledge overlay and chain distillation. All steps run on frozen embeddings without
fine-tuning.}
\label{fig:mech-pipeline}
\end{figure*}

\begin{figure}[p]
\centering
\includegraphics[width=\linewidth,height=0.9\textheight,keepaspectratio]{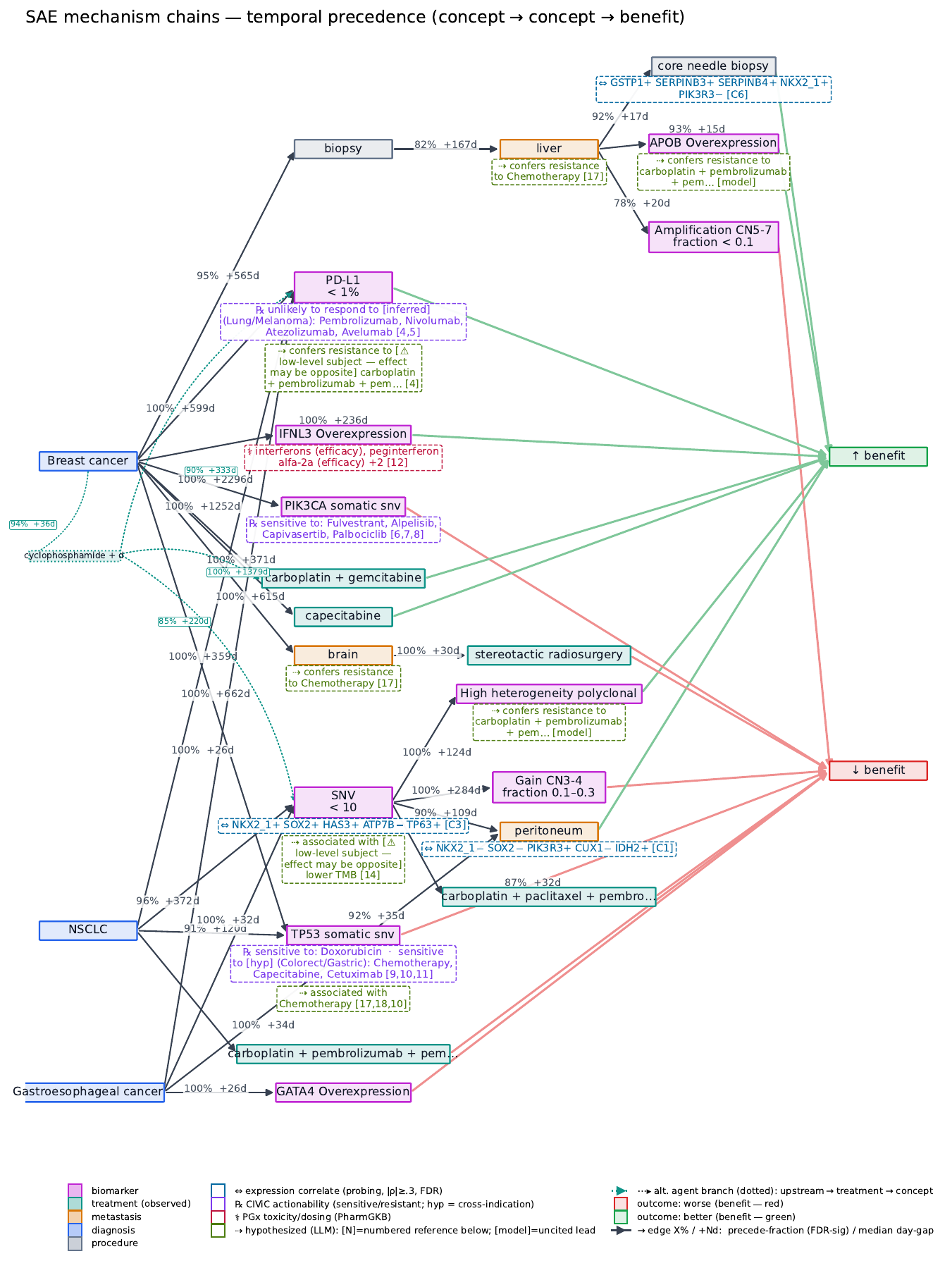}
\captionsetup{font=small}
\caption[Mechanism graph (illustrative cohort).]{\textbf{Mechanism graph.} Illustrative mechanism graph depicting feature interactions associated with treatment benefit in second-line or later (2L+) patients treated with trastuzumab deruxtecan (T-DXd) versus physician's choice chemotherapy in a multi-tumor cohort. Directed chains are distilled from the temporal-precedence directed acyclic graph (DAG). Concept nodes are color-coded by category (genomic biomarkers, treatment, diagnosis, procedure, metastatic site); temporal edges are labeled with precede-fraction and median day gap (in days); annotation chips mark curated gene-drug actionability (CIViC), RNA expression correlates derived via probing analysis, and LLM-hypothesized relations with their retrieved citations listed in Table~\ref{tab:mech-refs}.}
\label{fig:mech-pred-graph}
\end{figure}

\begin{enumerate}

\item \label{mech_stage:mech-attr} \textbf{Fact attribution:} Atomic facts (diagnoses, biomarkers, agents, lab values) are iteratively removed from a patient trajectory. We recompute the embedding and record the resulting score shift $\Delta s$. Within-patient normalization by total absolute attribution yields a signed, length-comparable metric for each fact.

\item \label{mech_stage:mech-sae} \textbf{Sparse feature:} A Sparse Autoencoder (SAE)~\citep{cunningham2023sparse} projects embedding $h$ into an overcomplete latent space $z$ under an $L_1$ sparsity constraint. The linear decoder yields sparse feature directions that facilitate downstream intervention in stage~\ref{mech_stage:mech-steer}, while enabling near-lossless score reconstruction ($\hat{h} \approx h$).

\item \label{mech_stage:mech-diff} \textbf{Differential activation testing and clustering:} Patients are partitioned by the population median of score $s(h)$. SAE latents are screened for differential activation between high- and low-score groups of patients using Mann--Whitney rank tests and Cohen's $d$, adjusted via Benjamini--Hochberg FDR control~\citep{benjamini1995controlling}. Co-activating latents are grouped into interpretable mechanisms via clustering with a hybrid Jaccard-cosine metric.

\item \label{mech_stage:mech-label} \textbf{Attribution-weighted concept labelling}: Mechanisms are labeled by scaling per-concept attribution strengths (see stage~\ref{mech_stage:mech-attr}) with cohort-wide inverse document frequency (IDF). This filters ubiquitous administrative events and highlights disproportionately attributed clinical concepts. Near-synonym concepts are normalized.

\item \label{mech_stage:mech-steer} \textbf{Causal steering:} To evaluate causality, mechanism latents are ablated (zeroed) or amplified (clamped to the 95th percentile) prior to decoding and re-scoring. Observed score shifts and class-assignment flips categorize mechanisms as drivers, suppressors, or amplifiable. All validated cluster latents are simultaneously ablated and re-scored, yielding a variance-explained statistic $\mathrm{VE} = 1 - \operatorname{Var}(s_{\text{ablated}})/\operatorname{Var}(s_{\text{original}})$. A coverage decomposition separates clustered from unclustered contributions and reports a concentration ratio (per-feature VE of clustered versus unclustered latents) to distinguish concentrated from diffuse mechanisms.

\item \label{mech_stage:mech-rag} \textbf{Retrieval-grounded mechanistic reasoning:} A local domain language model (MedGemma~4B~\citep{sellergren2026medgemma15technicalreport}) queries an offline FAISS index~\citep{douze2024faiss} containing ten public biomedical knowledge bases, namely CIViC~\citep{griffith2017civic}, OncoKB~\citep{chakravarty2017oncokb}, DrugBank~\citep{wishart2018drugbank} , PharmGKB~\citep{whirlcarrillo2021evidence}, Reactome~\citep{milacic2024reactome}, GO~\citep{go2023knowledgebase,ashburner2000gene}, MSigDB~\citep{liberzon2015molecular}, String~\citep{szklarczyk2023string}, PubMed~\citep{sayers2026ncbi}, and ClinicalTrials~\citep{zarin2011clinicaltrials},
embedded with BioLORD-2023~\citep{remy2024biolord}. The offline model outputs structured phenotype/hypothesis/evidence records and citation-grounded subject/relation/object triplets via retrieval-augmented generation~\citep{lewis2020retrieval},  supplying the hypothesized-edge layer of the mechanism graph (see stage~\ref{mech_stage:mech-graph}).

\item \label{mech_stage:mech-graph} \textbf{Graph Assembly: temporal precedence and knowledge overlay:}
To structure candidate concepts into a temporally ordered narrative, flat concept table entries are assembled into a directed acyclic graph (DAG; Figure~\ref{fig:mech-pred-graph}), connecting concepts to the outcome via statistically-tested precedence. Temporal edges ($A \to B$) are retained for concept pairs co-occurring in $\ge n_{\min}$ patients where $A$ precedes $B$ in $>60\%$ of cases (one-sided sign test, BH-FDR $q \le 0.05$). This temporal backbone is enriched with three multi-layer edge types: curated biological interactions (CIViC, STRING, MSigDB), external correlational associations ($|\rho| \ge 0.3$, FDR-controlled), and the RAG-generated triplets from stage~\ref{mech_stage:mech-rag}. Finally, the DAG is distilled into concise, domain-interpretable narrative chains (length 2-4 concepts to outcome) ranked by path weight and terminal concept diversity.

\end{enumerate}

The SAE acts as a discovery engine to identify, cluster, and test latent concepts, while the Mechanism Graph temporally orders these concepts across patient histories and links them to outcomes via established or novel candidate biological pathways. Designed as a generalizable mechanism discovery framework, the pipeline supports generating and evaluating biological hypotheses both for prognostic risk strata and for differential treatment benefit. 

To demonstrate its utility, we evaluated overall survival predictions in a multi-tumor cohort (metastatic breast cancer, non-small cell lung cancer [NSCLC], and gastroesophageal cancer) treated with trastuzumab deruxtecan (T-DXd) versus physician's choice chemotherapy in later-line settings. Reflecting real-world clinical practice and data completeness, patient eligibility was relaxed from HER2-positive status to include HER2-low, HER2-mutant, and HER2-unknown or unrecorded cases. In these cases, T-DXd demonstrates clinical activity via its potent bystander effect or is commonly prescribed in routine clinical practice. The pipeline recovered mechanisms consistent with established disease biology while generating plausible exploratory hypotheses for clinical follow-up (Figure~\ref{fig:mech-pred-graph}). Specifically, high-benefit outcomes are associated with features reflecting both prior antibody-drug conjugate (ADC) vulnerability and chemotherapy resistance. Across all three histologies, high benefit is associated with PD-L1-negative status ($<1\%$), a population that traditionally derives limited response from conventional immune checkpoint inhibitors. In the breast cancer subgroup, higher therapeutic benefit is associated with baseline core needle biopsy profiling and prior carboplatin/gemcitabine failure. Additionally, the pipeline surfaced metabolic signatures such as \textit{APOB} overexpression, which likely reflects site-specific liver biopsy context requiring further experimental isolation. Within the NSCLC cohort, high benefit correlated with pronounced intratumoral heterogeneity, a phenotype that could be consistent with sensitivity to the bystander killing effect of T-DXd. 

Conversely, low-benefit clusters are characterized by compromised drug delivery, reduced target antigen availability, or aggressive multi-drug resistance phenotypes. For example, breast cancer cases frequently presented with brain metastases, which may limit large-molecule delivery across the blood-brain barrier. Mechanistically, this low-benefit cluster aligns with a low copy-number variant (CNV) burden (Gain CN3-4 fraction $0.1$-$0.3$, Amplification CN5-7 fraction $<0.1$), indicating a reduced likelihood of the HER2 genomic alterations required for effective ADC target engagement. Furthermore, these NSCLC patients frequently had prior exposure to carboplatin/pembrolizumab/pemetrexed triple-therapy, which may reflect prior treatment selection for resistant tumor subclones. In the gastroesophageal cancer subset, the low-benefit phenotype was enriched for lineage-associated genomic features such as \textit{GATA4} overexpression; this may act as a tissue confounder rather than a response modifier, separating the two requires prospective work. Steering the feature clusters shifts the model's predicted risk score, so the mechanism graph provides an interpretable, directed framework mapping upstream clinical and genomic input events to predicted patient outcomes.

\begin{table}[htbp]
\centering
\small
\caption{Numbered references for the mechanism-chain (Figure~\ref{fig:mech-pred-graph}). Sources: PubMed entries in author/journal form; CIViC, PharmGKB and trial records in accession form. CIViC tiering includes cross-indication evidence.}
\label{tab:mech-refs}
\begin{tabular}{@{}r p{0.86\textwidth}@{}}
\toprule
\# & Reference \\
\midrule
1 & CIViC evidence EID993: KRAS Exon 2 Mutation in Colorectal Cancer (Predictive, level A). \\
2 & CIViC evidence EID2998: KRAS Mutation in Lung Non-small Cell Carcinoma (Predictive, level A). \\
3 & CIViC evidence EID5345: KRAS Mutation in Colorectal Cancer (Predictive, level A). \\
4 & CIViC evidence EID704: CD274 Expression in Melanoma (Predictive, level B). \\
5 & CIViC evidence EID1167: CD274 Expression in Lung Non-small Cell Carcinoma (Predictive, level B). \\
6 & CIViC evidence EID7313: PIK3CA Mutation in Breast Cancer (Predictive, level A). \\
7 & CIViC evidence EID12020: PIK3CA Mutation OR PTEN Mutation OR AKT1 Mutation in Breast Cancer (Predictive, level A). \\
8 & CIViC evidence EID12533: PIK3CA H1047R OR PIK3CA H1047Y OR PIK3CA H1047L OR PIK3CA H1047D OR PIK3CA H1047I OR PIK3CA H1047N OR PIK3CA H1047P OR PIK3CA H1047Q OR PIK3CA H1047T in Breast Cancer (Predictive, level A). \\
9 & CIViC evidence EID399: TP53 R249 in Breast Cancer (Predictive, level B). \\
10 & CIViC evidence EID850: TP53 Mutation in Gastric Adenocarcinoma (Predictive, level B). \\
11 & CIViC evidence EID875: TP53 Wildtype in Colorectal Cancer (Predictive, level B). \\
12 & PharmGKB: IFNL3–interferons;peginterferon alfa-2a;peginterferon alfa-2b;ribavirin (Efficacy, level 1B). \\
13 & Bellini MF et al. Alterations of the TP53 gene in gastric and esophageal carcinogenesis. J Biomed Biotechnol. 2012;2012:891961. PMID 22919278. \\
14 & Kumar D et al. Revisiting the Association of ECOG Performance Status With Clinical Outcomes in Diverse Patients With Cancer. J Natl Compr Canc Netw. 2024;22(2 D). PMID 38653321. \\
15 & Xing AY et al. p53 Missense Mutation is Associated with Immune Cell PD-L1 Expression in Triple-Negative Breast Cancer. Cancer Invest. 2022;40(10):879-888. PMID 35980253. \\
16 & Li X et al. Molecular Landscape of TP53/RB1 Co-Altered Tumors Uncovers Emerging Therapeutic Vulnerabilities. Genes Chromosomes Cancer. 2026;65(1):e70100. PMID 41562186. \\
17 & Kim T et al. Linking p53 immunostaining to TP53 mutation status in patients with non-small cell lung cancer. Pathology. 2025;57(7):881-889. PMID 40835511. \\
18 & Bharathiraja P et al. Solasodine targets NF-$\kappa$B signaling to overcome P-glycoprotein mediated multidrug resistance in cancer. Exp Cell Res. 2024;441(1):114153. PMID 39013486. \\
\bottomrule
\end{tabular}
\end{table}

\section{Conclusion}
We introduced the oFM, a multimodal foundation model that integrates longitudinal clinical history with patient-linked DNA and RNA findings and H\&E pathology embeddings. Across prognostic and comparative-treatment benchmarks, frozen oFM embeddings consistently outperformed strong baselines based on expert-curated clinical and molecular features, suggesting that the model captures information not readily recovered from conventional static features. The mechanism-discovery framework further provides an interpretable layer for downstream models built on oFM embeddings by linking their predictions to attributed clinical and molecular events, sparse latent features, and evidence-grounded temporal hypotheses. By jointly representing longitudinal clinical context with observed tumor biology, including DNA and RNA profiles, this framework enables downstream predictions to be interrogated in terms of biological and clinical factors associated with prognosis, treatment response, and resistance. These retrospective real-world evaluations are intended to compare representation quality rather than establish clinically validated biomarkers, treatment effects, or definitive biological mechanisms. External and prospective validation will be required, but the results support longitudinal multimodal modeling as a shared foundation for scalable, interpretable applications in oncology diagnostics and therapeutic development.

\section*{Acknowledgments}

We thank Arpita Saha, Gabriel Altay, Hongru Hu, Rohan Prakash Joshi, Sam Heilbroner, Sujoy Ganguly, Rafi Pelossof, Simona Cristea, Jiajie Xiao, and Fan Zhang for contributions to preliminary work underlying this project, including early knowledge transfer, development of the initial codebase and data-processing pipelines, and valuable discussions that informed the model training and evaluation protocols.

We thank Joe Oakley, MD, for valuable feedback and guidance on model interpretability and clinical use case formulation.

We thank Ryan Godart, Anirban Nandi, Daniel Cohn, and Evan Czyzycki for building and supporting the compute cluster that enabled large-scale model development and experimentation. We also thank Jonathan Wills, Maria Berezina, and Allison Maresca for organizing the data curation and delivery that enabled this work.

Finally, we thank Or Yaacov and Ben Terdich for contributions to the initial design of the benchmark task suite and for helping establish the benchmark evaluation protocol.

\bibliographystyle{plainnat}
\bibliography{refs}
\clearpage

\appendix
\section{Appendix}
\label{sec:appendix}

\setcounter{figure}{0}
\renewcommand{\thefigure}{A\arabic{figure}}
\setcounter{table}{0}
\renewcommand{\thetable}{A\arabic{table}}

\subsection{Data Sources}
\label{appendix:data_source}

To build the patient trajectories, we used data from the Tempus Data Model (TDM), a multimodal oncology data platform that integrates clinical and molecular data from multiple sources, as detailed below.

\subsubsection*{Clinical Data}
\begin{itemize}
    \item \textbf{Curated.} Structured clinical data manually abstracted by trained experts from patient medical records, including progress notes, pathology reports, and treatment plans. Abstraction followed standardized guidelines to capture diagnosis, treatment (including care plans, regimens, and lines of therapy), response assessments, disease progression, and other clinical endpoints.
    \item \textbf{Native.} Structured data from electronic health records (EHRs) transmitted directly from partner healthcare institutions, including medication administrations and orders, laboratory results, vital signs, encounters, performance status, and smoking status.
    \item \textbf{Pathology.} Diagnosis and histology information derived from Tempus pathologist review of biopsy specimens.
    \item \textbf{Third-party.} Supplementary data from external sources, including claims-based mortality data, used to augment vital status and last-known-alive dates.
    \item \textbf{Derived.} Algorithmically computed variables, including BMI calculated from height and weight measurements, and lines of therapy derived from curated care plan data.
\end{itemize}

In the case where a clinical concept is captured by both \textbf{curated} and \textbf{native} data (e.g., antineoplastic medications), records were harmonized to maximize completeness and date precision, with the source of each record tracked via a $\text{row\_source}$ indicator.

\subsubsection*{Molecular Data}
\begin{itemize}
    \item \textbf{Tempus-generated molecular data.} Results from Tempus next-generation sequencing (NGS) assays (xT, xF, xE, RS), including single nucleotide variants (SNVs), insertions/deletions (indels), copy number variants (CNVs), structural variants (SVs), RNA gene expression, microsatellite instability (MSI), tumor mutational burden (TMB), homologous recombination deficiency (HRD), HLA genotyping, immune infiltration, immune repertoire, neoepitope predictions, and microorganism detection.
    \item \textbf{Third-party reported molecular data.} Results from molecular tests performed by external laboratories, including NGS, PCR, immunohistochemistry (IHC), and fluorescence in situ hybridization (FISH), capturing biomarker results such as hormone receptor status (ER, PR, HER2), MSI, mismatch repair (MMR), and gene-level alterations.
    \item \textbf{Reference laboratory data}: Results from Tempus partner reference laboratories (e.g., minimal residual disease [MRD] testing, IHC/FISH).
\end{itemize}

\subsubsection*{De-identification}
All data were de-identified using HIPAA-compliant statistical de-identification (Expert Determination) methods. Dates were shifted uniformly per patient (within a range of $0-30$ days) to preserve temporal relationships between clinical events while protecting patient identity.

\subsection{Molecular blueprint}
\label{appendix:blueprint}

For each biospecimen, a molecular blueprint was assembled from DNA, RNA, IHC, and ISH, integrating ten categories:

\begin{itemize}
    \item Biospecimen metadata detailing histology and anatomical site,
    \item IHC biomarkers including PD-L1 TPS, minimal residual disease (MRD), and HER2 status,
    \item Genomic instability indicators encompassing MSI/dMMR status, tumor mutational burden ($\text{TMB-H} \ge 10\,\text{mutations/Mb}$), homologous recombination deficiency (HRD), HLA loss of heterozygosity (LOH), and clonal heterogeneity,
    \item Host immune descriptors comprising TIL phenotype, CD8/CD4 dominance, immune likelihood, and 20 HLA single-allele hotspots plus 9 multi-locus haplotypes across Class~I and II loci,
    \item High-confidence pathogen calls,
    \item Disruption in five canonical oncogenic signaling pathways (mTOR, TGF-$\beta$, RTK, HRD, DDR),
    \item Pathogenic or likely pathogenic SNVs/indels with complete genomic and protein annotations,
    \item Copy-number alterations and fraction of genome altered,
    \item Structural variants or fusions, and
    \item RNA-level expression abnormalities alongside 11 actionable splicing events (e.g., \emph{MET} exon~14 skipping, \emph{AR-V7}, \emph{EGFRvIII}).
\end{itemize}

\subsection{Prognostic classification: cohort stratification (tumor \texorpdfstring{$\times$}{x} therapy)}
\label{app:prognostic-cohort-stratification}

\paragraph{Cohort stratification.} Strata are defined as \texttt{general\_tumor\_type} $\times$ \texttt{therapy\_class\_group}, restricted to a whitelist of valid pairs; a stratum is dropped only if (a) it has zero embedded patients, or (b) it contains a single class (all positive or all negative). The reported metric is the unweighted macro-mean of per-stratum AUCs. This stratification tests whether performance is consistent across clinically comparable tumor--therapy combinations rather than being driven by a few high-prevalence cohorts. Because the macro-mean weights each scored stratum equally, results are sensitive to small strata; Table~\ref{tab:app-per-stratum-composition} provides the per-stratum sample sizes used for each task.

\begingroup
\setlength{\LTpre}{4pt}
\setlength{\LTpost}{4pt}
\footnotesize
\setlength{\tabcolsep}{3pt}
\renewcommand{\arraystretch}{1.10}

\begin{longtable}{
  >{\raggedright\arraybackslash}p{0.27\linewidth}
  >{\raggedright\arraybackslash}p{0.35\linewidth}
  >{\raggedleft\arraybackslash}p{0.105\linewidth}
  >{\raggedleft\arraybackslash}p{0.105\linewidth}
  >{\raggedleft\arraybackslash}p{0.105\linewidth}
}

\caption{Per-stratum composition by tumor type and therapy group, for the strata
entering the prognostic-classification analysis (minimum 20 patients per stratum, as
in the figure). Values are patients, $n$ (positive cases). An em dash indicates the
stratum did not enter that task: fewer than 20 patients, a single outcome class, or
no available embeddings.}
\label{tab:app-per-stratum-composition}\\

\toprule
\textbf{Tumor type} &
\textbf{Therapy group} &
\makecell[r]{\textbf{OS}\\$\boldsymbol{n}$ \textbf{(pos)}} &
\makecell[r]{\textbf{PFS}\\$\boldsymbol{n}$ \textbf{(pos)}} &
\makecell[r]{\textbf{Response}\\$\boldsymbol{n}$ \textbf{(pos)}} \\
\midrule
\endfirsthead

\multicolumn{5}{l}{
  \footnotesize\textit{Table \thetable\ continued from the previous page}
}\\
\toprule
\textbf{Tumor type} &
\textbf{Therapy group} &
\makecell[r]{\textbf{OS}\\$\boldsymbol{n}$ \textbf{(pos)}} &
\makecell[r]{\textbf{PFS}\\$\boldsymbol{n}$ \textbf{(pos)}} &
\makecell[r]{\textbf{Response}\\$\boldsymbol{n}$ \textbf{(pos)}} \\
\midrule
\endhead

\midrule
\multicolumn{5}{r}{
  \footnotesize\textit{Continued on the next page}
}\\
\endfoot

\bottomrule
\multicolumn{5}{
  >{\raggedright\arraybackslash}p{\dimexpr\linewidth-2\tabcolsep\relax}
}{
  \footnotesize\textit{Abbreviations:}
  ADC, antibody-drug conjugate;
  ICI, immune checkpoint inhibitor;
  IO, immuno-oncology;
  SM, small molecule;
  TKI, tyrosine kinase inhibitor.
}\\
\endlastfoot

Colorectal cancer
  & Biologic + chemotherapy
  & 383 (285) & 355 (280) & 451 (196)\\

NSCLC
  & Chemotherapy + IO/ICI
  & 325 (207) & 327 (214) & 450 (181)\\

Breast cancer
  & Hormone + targeted (SM)
  & 239 (203) & 279 (214) & 277 (94)\\

Colorectal cancer
  & Chemotherapy
  & 282 (226) & 248 (199) & 240 (100)\\

Pancreatic cancer
  & Chemotherapy
  & 274 (148) & 205 (122) & 281 (62)\\

Prostate cancer
  & Hormone
  & 299 (290) & 326 (287) & 100 (44)\\

Ovarian/fallopian tube/peritoneal cancer
  & Chemotherapy
  & 201 (193) & 199 (189) & 114 (79)\\

NSCLC
  & IO/ICI
  & 114 (67) & 103 (68) & 113 (41)\\

NSCLC
  & TKI
  & 99 (74) & 102 (74) & 115 (50)\\

Tumor of unknown origin
  & Chemotherapy
  & 114 (63) & 98 (56) & 83 (29)\\

Breast cancer
  & Chemotherapy
  & 100 (65) & 96 (55) & 92 (31)\\

Melanoma
  & IO/ICI
  & 82 (61) & 94 (71) & 90 (32)\\

Gastroesophageal cancer
  & Chemotherapy
  & 96 (39) & 63 (37) & 97 (32)\\

NSCLC
  & Chemotherapy
  & 107 (68) & 83 (57) & 65 (23)\\

Prostate cancer
  & Hormone + other
  & 88 (81) & 95 (84) & 55 (27)\\

Gastroesophageal cancer
  & Chemotherapy + IO/ICI
  & 67 (44) & 65 (41) & 84 (27)\\

Tumor of unknown origin
  & Chemotherapy + IO/ICI
  & 60 (32) & 63 (34) & 78 (23)\\

Prostate cancer
  & Chemotherapy + hormone
  & 64 (59) & 71 (60) & 52 (25)\\

Ovarian/fallopian tube/peritoneal cancer
  & Chemotherapy + targeted (SM)
  & 49 (45) & 61 (52) & 66 (50)\\

Breast cancer
  & Hormone
  & 66 (52) & 62 (49) & 31 (16)\\

Urinary tract cancer
  & Chemotherapy
  & 52 (29) & 49 (25) & 53 (20)\\

Endometrial cancer
  & Chemotherapy
  & 54 (41) & 54 (45) & 39 (25)\\

SCLC
  & Chemotherapy + IO/ICI
  & 37 (19) & 30 (13) & 77 (31)\\

Kidney cancer
  & IO/ICI
  & 42 (33) & 52 (39) & 48 (13)\\

Ovarian/fallopian tube/peritoneal cancer
  & Biologic + chemotherapy
  & 51 (44) & 50 (47) & 41 (24)\\

Lung cancer
  & Chemotherapy + IO/ICI
  & 36 (14) & 35 (18) & 53 (15)\\

Kidney cancer
  & IO/ICI + TKI
  & 35 (28) & 39 (29) & 48 (19)\\

Breast cancer
  & Biologic + chemotherapy
  & 32 (28) & 42 (35) & 36 (20)\\

Breast cancer
  & Chemotherapy + IO/ICI
  & 29 (20) & 31 (24) & 44 (12)\\

Colorectal cancer
  & IO/ICI
  & 36 (30) & 35 (30) & 32 (16)\\

Urinary tract cancer
  & ADC + IO/ICI
  & 34 (27) & 33 (29) & 33 (14)\\

Gallbladder/biliary tract cancer
  & Chemotherapy
  & 27 (12) & 25 (13) & 43 (12)\\

Gallbladder/biliary tract cancer
  & Chemotherapy + IO/ICI
  & 34 (25) & 28 (18) & 29 (6)\\

Endometrial cancer
  & Chemotherapy + IO/ICI
  & 25 (18) & 29 (24) & 31 (14)\\

Urinary tract cancer
  & IO/ICI
  & 26 (14) & 24 (13) & 31 (8)\\

Gastroesophageal cancer
  & Biologic + chemotherapy
  & 23 (16) & 21 (16) & 27 (10)\\

Gastroesophageal cancer
  & Biologic + chemotherapy + IO/ICI
  & 20 (14) & 22 (15) & 29 (11)\\

Head and neck SCC
  & Chemotherapy + IO/ICI
  & 24 (13) & \textemdash & 37 (6)\\

Prostate cancer
  & Chemotherapy + hormone + other
  & 20 (19) & 23 (20) & \textemdash\\

Urinary tract cancer
  & Chemotherapy + IO/ICI
  & \textemdash & \textemdash & 39 (11)\\

Ovarian/fallopian tube/peritoneal cancer
  & Biologic + chemotherapy + targeted (SM)
  & \textemdash & \textemdash & 21 (15)\\

\end{longtable}
\endgroup

\subsection{Baseline Feature Selection}
\label{appendix:feature-selection}

7520 candidate features were drawn from a versioned catalogue of standardized, category-annotated clinical and molecular features. Treatment intent labels (adjuvant, neoadjuvant, induction, consolidation, conditioning, and maintenance) were excluded across all configurations to prevent treatment assignment leakage. Feature selection was performed independently for each cohort/task using only the training and validation data; test data were strictly withheld.

Candidate features passed through three successive selection stages: 

\begin{enumerate}
\item Variance filtering: Features with zero variance or falling below the 70th percentile of candidate variance were removed. 
\item Univariate screening: Remaining candidates were evaluated in individual penalized Cox proportional-hazards models (penalty = 0.01, elastic-net mixing = 0.9) incorporating delayed entry. Features were retained if the concordance index exceeded 0.5 and the Wald p-value was below 0.05. For cohorts exceeding 10,000 patients, median concordance indices and p-values across 10 fixed-seed subsamples of 10,000 patients were used to stabilize estimates. 
\item Multivariate selection: Retained candidates were fit jointly in a single penalized Cox model; features with absolute coefficients below 0.01 were discarded.
\end{enumerate}

A core set of base clinical features was retained across all baselines regardless of statistical significance, subject only to variance filtering, to preserve cross-cohort comparability: ECOG performance status (0--5), four age bands, female sex, tumor grade (1--4), and stage (0--IV). Missing or constant levels within a cohort were dropped. Continuous variables were normalized with canonical standardization (mean, standard deviation on the training set) and were imputed to zero. Combining the selected candidate features with these base features yields the final compact, cohort-specific feature set. Tab.~\ref{tab:num_baseline_features_selected} shows the number of features selected per predictive cohort.

\begin{table}[h]
    \centering
    \begin{tabular}{@{}>{\raggedright\arraybackslash}p{0.8\linewidth}r@{}}
\toprule
Cohort & Selected features \\
\midrule
Breast HR+/HER2$-$ CDK4/6 and aromatase inhibitor (AI) comb.\ vs.\ AI alone, 1L & 104 \\
CRC RAS WT anti-EGFR vs.\ anti-VEGF, 1L                                        & 197 \\
NSCLC pembro--chemo combo vs.\ chemo, 1L                                       &  94 \\
NSCLC pembro--chemo combo vs.\ pembro alone, 1L                                &  90 \\
NSCLC dual IO vs.\ IO--chemo combo, 1L                                         &  81 \\
mRCC dual IO vs.\ IO--TKI, 1L                                                  & 160 \\
HSPC ADT doublet vs.\ triplet, 1L                                              & 114 \\
Enriched multi-tumor Enhertu vs.\ chemo, 2L+                                   & 162 \\
Pancancer Enhertu vs.\ chemo, 2L+                                              & 149 \\
Pancancer IO--chemo vs.\ IO alone, 1L                                          & 165 \\
Pancancer HRR+ PARPi vs.\ chemo, 2L+                                           & 111 \\
\bottomrule
\end{tabular}
    \caption{Number of selected baseline features per predictive evaluation cohort.}
    \label{tab:num_baseline_features_selected}
\end{table}

\subsection{Inverse-Propensity Weighting}
\label{sec:uv-ipw}

We use stabilized inverse-propensity weighting to correct treatment selection bias in the observational cohorts~\citep{Robins2000,Austin2015}. The propensity model is trained on a set of always-included features and a subset of the features selected as described in Appendix~\ref{appendix:feature-selection}. The always-included features are one-hot ECOG performance status ($0$--$5$), age band ($\leq$30, 30--50, 50--75, $>$75), gender, tumor grade ($1$--$4$), and tumor stage ($0$--$4$). Selected features may include IHC biomarkers, tumor type, surgical pathology outcomes, smoking status, and biopsy vs resection. Treatment information is dropped. A logistic model on this feature set estimates the propensity $e(\mathbf{x}_i) = \hat{P}(W_i = 1 \mid \mathbf{x}_i)$, clipped to $[0.01, 0.99]$, and the stabilized weights \citep{Xu2010} are calculated from propensity scores

\begin{equation}
w_i =
\begin{cases}
\hat{P}(W=1)\,/\,e(\mathbf{x}_i) & W_i = 1,\\[2pt]
\hat{P}(W=0)\,/\,\bigl(1-e(\mathbf{x}_i)\bigr) & W_i = 0,
\end{cases}
\end{equation}

where $\hat{P}(W=1)$ is the treated fraction of the cohort and $\hat{P}(W=0) = 1 - \hat{P}(W=1)$.
These weights enter the Cox partial log-likelihood when training the UV-Cox estimator (Appendix~\ref{sec:uv-optimization}). The propensity scores are reused with augmented inverse-propensity weighting (AIPW) during evaluation~\citep{robins1994estimation}.

\subsection{Two-Stage UV-Cox Predictive Evaluation Design}
\label{sec:uv-optimization}

To evaluate treatment benefit in our benchmarks, we set up and fit a convex, linear Cox~\citep{Cox1972} model that represents the residual treatment-dependent risk beyond treatment-independent prognostic risk. Let $W_i \in \{0,1\}$ be the treatment assignment. For $\mathbf{x}_i \in \R^{p}$, a representation of patient $i$ (the oFM embedding $h(t^*)$, or the selected features for the baseline),  we split the linear predictor $\eta_i$ into a prognostic term and a treatment-modifying term:

\begin{equation}
\begin{aligned}
\eta_i &= u(\mathbf{x}_i) + W_i\, v(\mathbf{x}_i), \\
\lambda(t \mid \mathbf{x}_i, W_i) &= \lambda_0(t)\,\exp\!\bigl(u(\mathbf{x}_i) + W_i\, v(\mathbf{x}_i)\bigr),
\end{aligned}
\end{equation}

where $\lambda$ measures patient risk and $\lambda_0$ defines baseline risk. Here $u$ is the prognostic head which carries the risk present in both arms and $v$ is the predictive head which carries the residual treatment-dependent risk. The treatment-benefit score is $\tau(\mathbf{x}_i) = -v(\mathbf{x}_i)$, so $v(\mathbf{x}_i) < 0$ predicts a lower hazard under treatment. Both heads are linear with feature-weights $\beta$, and $b_v$ modeling the main treatment effect in the treatment head:

\begin{equation}
\begin{aligned}
u(\mathbf{x}_i) &= \bm{\beta}_u^{\top}\tilde{\mathbf{x}}_i, \\
v(\mathbf{x}_i) &= \bm{\beta}_v^{\top}\tilde{\mathbf{x}}_i + b_v.
\end{aligned}
\end{equation}

Features are z-scored with moments computed on training and validation data, not the test set.  Both heads are fit by minimizing a weighted negative log partial likelihood with L2 regularization,

\begin{equation}
\ell(\bm{\eta}) =
-\frac{1}{\sum_{j:\delta_j=1} w_j}
\sum_{i:\delta_i=1} w_i\!\left[\eta_i - \log\!\!\sum_{j \in R(T_i)} w_j\, e^{\eta_j}\right],
\qquad
R(t) = \{\, j : L_j < t \le T_j \,\},
\label{eq:coxpl}
\end{equation}

where $\delta_i \in \{0,1\}$ is the event indicator and the risk set $R(t)$ enforces delayed entry, so a patient contributes only after the left-truncation entry time $L_i$ \citep{Klein2003}. To avoid leaking prognostic signal into the predictive head, fitting is done in two stages. With the predictive head $v$ is initialized to zero, stage 1 fits only the prognostic head $u$. Stage 2 then freezes the prognostic head $u$ and fits the predictive head $v$.

Each stage is optimized with full-batch L-BFGS. The L2 penalty is selected with 5-fold cross-validation over the union of the training and validation sets. In stage 1, L2 is selected over a 22-point logarithmic grid on $[10^{-4}, 10^{3}]$. The model is refit on the full train${}+{}$validation set at the selected L2 penalty. In stage 2, the same L2 penalty is used, as lowering it risks leaking prognostic signal into $v$ and raising it shrinks $v$, reducing treatment-dependent signal.

\subsection{Evaluation Metrics}
\label{sec:uv-metrics}

We evaluate predictive performance of oFM embeddings or baseline features with AUTOC and HRR and prognostic performance with C-index.

\paragraph{AUTOC.}
When ranking patients by predicted treatment benefit, the area under the targeting operator characteristic (AUTOC) is the average additional benefit compared to the cohort as a whole, computed over all possible top percentiles of patients. It is zero when the ranking carries no information about who benefits. The treatment benefit score $-v(\mathbf{x}_i)$ is estimated by the UV-Cox method (see Appendix~\ref{sec:uv-optimization}). However, ground truth benefit is never observed per patient as patients only ever occur in one arm and observation is right-censored. To estimate it, a pseudo-outcome $\theta_i$ is first computed per-patient using the restricted mean survival time (RMST) which is the average event-free time within a horizon $L$~\citep{Royston2013,Andersen2004}. We set $L$ to the $80$th percentile of observed follow-up, or the administrative censoring time where one is defined. Each patient's ground truth treatment benefit $\psi_i$ is then estimated from the pseudo-outcome $\theta_i$ with a doubly robust AIPW score \citep{robins1994estimation}, combining the propensity $e$ of Appendix~\ref{sec:uv-ipw} with arm-specific ridge outcome models fit on train${}+{}$validation and applied to the test patients. Ranking patients by estimated benefit, the targeting operator characteristic (TOC) and its integral are

\begin{equation}
\mathrm{TOC}(q) = \frac{1}{\lceil q n \rceil}\sum_{i\,\in\,\text{top-}q} \psi_i
                 \;-\; \frac{1}{n}\sum_{i=1}^{n} \psi_i,
\qquad
\mathrm{AUTOC} = \int_0^1 \mathrm{TOC}(q)\, dq,
\end{equation}

which weights the ranking uniformly \citep{Yadlowsky2025RATE}.

AUTOC is measured in time and the scale of AUTOC depends on the RMST and observational horizon of each cohort. We normalize AUTOC by SD where SD is the unweighted standard deviation of $\theta$, representing the relative scale of each cohort. Statistical significance is computed against a permutation null that randomly reshuffles treatment assignment within propensity-score bins; the $p$-value is the fraction of replicates whose AUTOC meets or exceeds the observed value.

\paragraph{Hazard ratio and hazard-ratio ratio.}
Whereas a hazard ratio (HR) measures how much treatment helps a group of patients, the hazard-ratio ratio (HRR) divides the hazard ratio of the estimated high-benefit patients by that of the estimated low-benefit patients. An HRR below one indicates correct separation according to treatment effect and an HRR above one indicates the ranking is inverted \citep{Iasonos2016}. Test patients are split into low- and high-benefit groups at the median benefit score, and a newly fit Cox model on the treatment indicator alone gives each stratum's $\mathrm{HR}$ using the IPW of Appendix~\ref{sec:uv-ipw}. Statistical significance is computed by a two-sided Wald test of $\log\mathrm{HRR} = 0$.

\paragraph{Prognostic discrimination.}
We report Uno's IPCW concordance index (C-index) \citep{Uno2011} of the prognostic score $u(\mathbf{x}_i)$. Across all patient pairs, it weights each pair to correct for right censoring and keeps only those that delayed entry leaves comparable. It is computed within each treatment arm so that it reflects baseline risk rather than treatment effect. Unlike in the predictive setting, the prognostic-only model is  fit on both arms pooled without IPW, which is needed only in the predictive setting to correct treatment selection bias. The C-index is then computed within each arm so that it reflects baseline risk rather than treatment effect. C-index 0.5 and 1 indicate random chance and correct ordering, respectively. The 95\% bootstrap confidence interval is computed over 1,000 iterations.

\paragraph{Pooling across cohorts.}
Per-cohort effects $y_i$ with standard errors $e_i$ are combined by Paule-Mandel random-effects meta-analysis \citep{PauleMandel1982}, which weights cohort $i$ by $1/(e_i^2 + \tau^2)$ for a between-cohort variance $\tau^2$ estimated from the data, so noisier cohorts are weighted less. $\mu_{\mathrm{AUTOC}}$ is the pooled AUTOC/SD. Dividing it by its standard error $\mathrm{SE}_{\mathrm{AUTOC}}$ gives $t_{\mathrm{AUTOC/SD}}$, which measures both effect size (since $\mathrm{SE}_{\mathrm{AUTOC}}$ grows with per-cohort noise) and consistency across cohorts (since $\mathrm{SE}_{\mathrm{AUTOC}}$ grows with disagreement between cohorts). HRR is pooled on the log scale and transformed back and C-index is pooled on logits. The significance of pooled AUTOC/SD and HRR results is tested against their permutation nulls where the per-patient treatment arm assignment (i.e. control v.s. treatment.) is randomly permuted. The statistical significance of the difference between the oFM and baseline pooled C-index values is computed with a Student t-test.

\subsection{Survival benchmark cohorts}
\begingroup
\footnotesize
\setlength{\tabcolsep}{5pt}
\setlength{\LTpre}{2pt}
\setlength{\LTpost}{2pt}
\renewcommand{\arraystretch}{1.05}
\begin{longtable}{@{} >{\raggedright\arraybackslash}p{0.19\linewidth} >{\raggedright\arraybackslash}p{0.25\linewidth} >{\raggedright\arraybackslash}p{0.28\linewidth} >{\raggedleft\arraybackslash}p{0.095\linewidth} >{\raggedleft\arraybackslash}p{0.095\linewidth} @{}}
\caption{\textbf{Survival-benchmark cohorts.} Two-arm comparative cohorts evaluated for differential treatment benefit on overall survival. Each row gives the cohort (with line-of-therapy and clinical-use-case tags), the experimental vs.\ control regimens, the clinical question (condensed to its primary clause), and the number of experimental~/~control patients in the combined train+validation (model-development) and held-out test splits. Patient counts are for the genomic (DNA+RNA) subset.}
\label{tab:cohorts-pred} \\
\toprule
\textbf{Cohort} & \textbf{Experimental / Control arm} & \textbf{Clinical question} & \makecell[r]{\textbf{Train+val}\\\textbf{(exp\,/\,ctrl)}} & \makecell[r]{\textbf{Test}\\\textbf{(exp\,/\,ctrl)}} \\
\midrule
\endfirsthead
\multicolumn{5}{l}{\tablename\ \thetable{} -- \textit{continued from previous page}} \\
\toprule
\textbf{Cohort} & \textbf{Experimental / Control arm} & \textbf{Clinical question} & \makecell[r]{\textbf{Train+val}\\\textbf{(exp\,/\,ctrl)}} & \makecell[r]{\textbf{Test}\\\textbf{(exp\,/\,ctrl)}} \\
\midrule
\endhead
\midrule
\multicolumn{5}{r}{\textit{Continued on next page}} \\
\endfoot
\bottomrule
\endlastfoot
\textbf{Breast HR+/HER2- CDK4/6 and Aromatase inhibitors (AI) comb vs AI alone 1L}
  & \textbf{Exp:} CDK4/6 inhibitor plus aromatase inhibitor (AI) \newline \textbf{Ctrl:} aromatase inhibitor alone
  & Does adding a CDK4/6 inhibitor (ribociclib, palbociclib, or abemaciclib) to first-line aromatase inhibitor therapy improve progression-free and overall survival in HR+/HER2- metastatic breast cancer?
  & 814\,/\,146 & 291\,/\,63 \\
\midrule
\textbf{CRC RAS WT anti-EGFR vs anti-VEGF 1L}
  & \textbf{Exp:} anti-EGFR antibody (cetuximab or panitumumab) combined with FOLFOX or FOLFIRI chemotherapy \newline \textbf{Ctrl:} bevacizumab combined with FOLFOX
  & In RAS wild-type metastatic colorectal cancer, does first-line anti-EGFR antibody plus chemotherapy improve overall survival versus bevacizumab plus chemotherapy?
  & 204\,/\,427 & 72\,/\,144 \\
\midrule
\textbf{NSCLC Pembro Chemo combo vs Chemo 1L}
  & \textbf{Exp:} pembrolizumab plus platinum (carboplatin or cisplatin) plus pemetrexed \newline \textbf{Ctrl:} platinum plus pemetrexed (chemotherapy only)
  & In driver-negative metastatic NSCLC, does adding pembrolizumab to platinum plus pemetrexed improve overall survival versus chemotherapy alone, and across PD-L1 subgroups ($<$1\%, 1--49\%, $\geq$50\%), which patients derive the greatest incremental benefit?
  & 1034\,/\,113 & 323\,/\,30 \\
\midrule
\textbf{NSCLC Pembro Chemo combo vs Pembro alone 1L}
  & \textbf{Exp:} pembrolizumab plus platinum-based chemotherapy \newline \textbf{Ctrl:} pembrolizumab monotherapy
  & In PD-L1 $\geq$50\% driver-negative metastatic NSCLC, does adding chemotherapy to first-line pembrolizumab improve outcomes versus pembrolizumab monotherapy?
  & 259\,/\,311 & 87\,/\,101 \\
\midrule
\textbf{NSCLC dual IO vs IO Chemo combo 1L}
  & \textbf{Exp:} chemo-immunotherapy with pembrolizumab plus platinum plus pemetrexed or paclitaxel (KEYNOTE-189 style) \newline \textbf{Ctrl:} dual checkpoint immunotherapy with ipilimumab plus nivolumab, with or without chemotherapy (CheckMate-227 style)
  & In driver-negative metastatic NSCLC, does first-line chemo-immunotherapy (pembrolizumab-based) outperform dual checkpoint blockade (ipilimumab plus nivolumab)?
  & 1231\,/\,111 & 398\,/\,54 \\
\midrule
\textbf{mRCC dual IO vs IO-TKI 1L}
  & \textbf{Exp:} IO-TKI combinations (pembrolizumab plus axitinib, lenvatinib plus pembrolizumab, cabozantinib plus nivolumab, or avelumab plus axitinib) \newline \textbf{Ctrl:} dual checkpoint immunotherapy (ipilimumab plus nivolumab)
  & In metastatic renal cell carcinoma, does first-line IO-TKI combination therapy improve outcomes versus dual checkpoint immunotherapy (ipilimumab plus nivolumab)?
  & 269\,/\,191 & 81\,/\,74 \\
\midrule
\textbf{HSPC ADT doublet vs triplet 1L}
  & \textbf{Exp:} ADT plus androgen receptor pathway inhibitor (ARPi) plus docetaxel (triplet therapy) \newline \textbf{Ctrl:} ADT plus ARPi (doublet therapy)
  & In metastatic hormone-sensitive prostate cancer, does adding docetaxel to ADT plus ARPi (triplet intensification) improve overall survival versus ADT plus ARPi alone (doublet)?
  & 278\,/\,1180 & 105\,/\,373 \\
\midrule
\textbf{Enriched multi-tumor Enhertu vs Chemo 2L+}
  & \textbf{Exp:} T-DXd (Enhertu) \newline \textbf{Ctrl:} standard chemotherapy
  & Across breast, gastric, and NSCLC cancers previously treated with trastuzumab and a taxane, does T-DXd improve progression-free and overall survival versus standard chemotherapy?
  & 231\,/\,447 & 55\,/\,148 \\
\midrule
\textbf{Pancancer Enhertu vs Chemo 2L+}
  & \textbf{Exp:} T-DXd (Enhertu / trastuzumab deruxtecan) \newline \textbf{Ctrl:} standard chemotherapy (irinotecan, capecitabine, eribulin, gemcitabine, or paclitaxel)
  & Across metastatic cancers previously treated with trastuzumab and a taxane, does T-DXd improve progression-free and overall survival versus standard chemotherapy?
  & 310\,/\,2125 & 82\,/\,702 \\
\midrule
\textbf{Pancancer IO Chemo vs IO alone 1L}
  & \textbf{Exp:} immune checkpoint inhibitor (ICI) plus chemotherapy \newline \textbf{Ctrl:} ICI alone
  & Across metastatic cancers, does adding chemotherapy to first-line immune checkpoint inhibitor therapy improve outcomes versus ICI monotherapy?
  & 3620\,/\,1877 & 1229\,/\,657 \\
\midrule
\textbf{Pancancer HRR+ PARPi vs Chemo 2L+}
  & \textbf{Exp:} PARP inhibitor (olaparib, niraparib, rucaparib, or talazoparib) \newline \textbf{Ctrl:} chemotherapy (docetaxel, paclitaxel, or platinum)
  & Across metastatic cancers with a homologous-recombination-repair (HRR) mutation or HRD, second-line or later, does a PARP inhibitor improve overall survival versus chemotherapy?
  & 224\,/\,532 & 82\,/\,200 \\
\end{longtable}
\endgroup

\subsection{Model parameter counts}

\begin{table}[h]
    \centering
    \begin{tabular}{lr}
    \toprule
    Component & Parameters \\
    \midrule
    \multicolumn{2}{l}{\emph{Trained parameters: oFM}} \\
    Episode encoder (GatorTron-base-2k + FoNE)      & 358M \\
    \quad transformer layers                        & \quad 303M \\
    \quad token-embedding matrix ($50{,}102\times1024$) & \quad 51.3M \\
    \quad position/type embeddings + LayerNorm      & \quad 2.1M \\
    \quad FoNE numeric embedder                     & \quad 1.1M \\
    \addlinespace
    Trajectory encoder (2-layer, RoPE)              & 25.2M \\
    \addlinespace
    Heads and projectors                            & 38.1M \\
    \quad predictor (FiLM residual MLP)             & \quad 10.5M \\
    \quad pathology projector ($8192\to1024$)       & \quad 8.0M \\
    \quad pathology reconstruction head             & \quad 8.4M\\
    \quad episode reconstruction head               & \quad 8.4M \\
    \quad time embedding                            & \quad 1.3M \\
    \quad survival head                             & \quad 1.0M \\
    \midrule
    Total trained                                   & \textbf{421M} \\
    \midrule
    \multicolumn{2}{l}{\emph{Frozen parameters: pathology encoder}} \\
    PRISM2 slide encoder          & 5.1B \\
    \quad Virchow2 tile encoder                & 632M \\
    \quad Perceiver + attention pooler (base/prognostic slide encoder) & 620M \\
    \quad Phi-3-Mini (diagnostic slide encoder) & 3.8B \\
    \bottomrule
    \end{tabular}
    \caption{Parameter counts for different components in the oFM model.}
    \label{tab:model_parameter_count}
\end{table}

\end{document}